\documentclass[preprint,10pt]{elsarticle}

\usepackage{amsmath,amssymb,amsfonts}%
\usepackage{amsthm}
\usepackage{multirow}
\usepackage{subfigure}
\usepackage{booktabs}
\usepackage{color}
\usepackage{hyperref}
\usepackage{enumitem}
\usepackage{xurl}

\journal{Computer Methods and Programs in Biomedicine}

\begin{document}

\begin{frontmatter}



\title{Early Detection of Multidrug Resistance Using Multivariate Time Series Analysis and Interpretable Patient-Similarity Representations}


\author[inst1]{Óscar Escudero-Arnanz\corref{cor1}}
\ead{oscar.escudero@urjc.es}
\author[inst1]{Antonio G. Marques}
\ead{antonio.garcia.marques@urjc.es}
\author[inst1]{Inmaculada Mora-Jiménez}
\ead{inmaculada.mora@urjc.es}
\author[inst2]{Joaquín Álvarez-Rodríguez}
\ead{joaquin.alvarez@salud.madrid.org}
\author[inst1]{Cristina Soguero-Ruiz}
\ead{cristina.soguero@urjc.es}
\cortext[cor1]{Corresponding author}

\affiliation[inst1]{organization={Department of Signal Theory and Communications, King Juan Carlos University},
            addressline={Camino del Molino, 5}, 
            city={Fuenlabrada},
            postcode={28942}, 
            state={Madrid},
            country={Spain}}

\affiliation[inst2]{organization={University Hospital of Fuenlabrada},
            addressline={Camino del Molino, 2}, 
            city={Fuenlabrada},
            postcode={28942}, 
            state={Madrid},
            country={Spain}}

\begin{abstract} 



\textbf{Background and Objectives}: 
\textcolor{black}{Multidrug Resistance (MDR) has been identified by the World Health Organization as a major global health threat}. It leads to severe social and economic consequences, including extended hospital stays, increased healthcare costs, and higher mortality rates.   \textcolor{black}{In response to this challenge, this study proposes a novel interpretable Machine Learning (ML) approach for predicting MDR, developed with two primary objectives: accurate inference and enhanced explainability.}

\textbf{Methods}: \textcolor{black}{\textit{For inference}, the proposed method is based on patient-to-patient similarity representations to predict MDR outcomes. Each patient is modeled as a Multivariate Time Series (MTS), capturing both clinical progression and interactions with similar patients. To quantify these relationships, we employ MTS-based similarity metrics, including feature engineering using descriptive statistics, Dynamic Time Warping, and the Time Cluster Kernel. These methods are used as} inputs for MDR classification through Logistic Regression, Random Forest, and Support Vector Machines, with dimensionality reduction and kernel transformations applied to enhance model performance. \textit{For explainability}, we employ graph-based methods to extract meaningful patterns from the data. Patient similarity \textcolor{black}{networks are generated using the MTS-based similarity metrics mentioned above}, while spectral clustering and t-SNE are applied to identify MDR-related subgroups, uncover clinically relevant patterns, and visualize high-risk clusters. These insights improve interpretability and support more informed decision-making in critical care settings.

\textbf{Results}: We validate our architecture on real-world Electronic Health Records from the \textcolor{black}{Intensive Care Unit (ICU)} dataset at the University Hospital of Fuenlabrada, achieving a \textcolor{black}{Receiver
Operating Characteristic Area Under the Curve} of 81\%. \textcolor{black}{Our framework surpasses ML and deep learning models on the same dataset by leveraging graph-based patient similarity. In addition, it offers a simple yet effective interpretability mechanism that facilitates the identification of key risk factors—such as prolonged antibiotic exposure, invasive procedures, co-infections, and extended ICU stays—and the discovery of clinically meaningful patient clusters.} For transparency, all results and code are available at~{\small \url{https://github.com/oscarescuderoarnanz/DM4MTS}}.

\textbf{Conclusions}: This study demonstrates the effectiveness of patient similarity representations and graph-based methods for MDR prediction and interpretability. The approach enhances prediction, identifies key risk factors, and improves patient stratification, enabling early detection and targeted interventions, highlighting the potential of interpretable ML in critical care.

\end{abstract}




\end{frontmatter}


\section{Introduction}
\label{sec:introduction}

Antimicrobial Resistance (AMR) occurs when pathogens develop the ability to withstand the effects of antimicrobial agents that were once effective against them~\cite{whoAntimicrobialResistance_2023}. Identified by the World Health Organization (WHO) as one of the most pressing global health threats, AMR has severe consequences for both public health and the global economy. By 2050, it is projected that AMR could lead to 10 million deaths annually, along with additional healthcare costs amounting to $1$ trillion US dollars per year and an estimated $3.4$ trillion US dollars in global GDP losses~\cite{whoAntimicrobialResistance_2023, AMR_deaths_2023}. 
Of particular concern within the broader context of AMR is Multidrug Resistance (MDR), a phenomenon where pathogens develop resistance to multiple antimicrobial agents, making infections significantly more difficult to treat. \textcolor{black}{Specifically, MDR is typically defined as resistance to at least one agent in three or more antimicrobial classes~\cite{magiorakos2012multidrug}}. The MDR problem is especially severe in high-risk settings like Intensive Care Units (ICUs), where vulnerable patients are more likely to become infected, and MDR bacteria can spread rapidly. Addressing MDR is crucial due to the reduction in treatment options, increased mortality rates, and prolonged hospital stays, all of which contribute to drastically elevated healthcare costs~\cite{world2024bacterial}. 
A further challenge in treating MDR infections lies in the delays associated with traditional diagnostic methods. Current approaches \textcolor{black}{for} determining whether a patient has developed resistance to multiple drugs typically take between 48 to 72 hours, as they rely on the time-consuming process of culturing pathogens. This diagnostic delay can undermine treatment efficacy and worsen clinical outcomes~\cite{tang2022machine}.
 
In this context, the vast availability of clinical data, particularly from Electronic Health Records (EHRs), offers significant opportunities for early diagnosis, enhanced medical care, and informed clinical decision-making~\cite{xie2022deep}. However, this data often presents challenges, as it is frequently incomplete, heterogeneous (EHRs combine information from diverse sources and devices), time-varying, and irregularly sampled, with varying frequencies and sequence lengths~\cite{moskovitch2022multivariate, martin2024irregular}.
These issues highlight the importance of having high-quality data, which refers to datasets that are complete, consistent, and well-structured, effectively addressing challenges such as missing values, noise, and inconsistencies~\cite{escudero2024low}. Without such quality, even large datasets may fail to support the training of robust Artificial Intelligence (AI) models. Addressing these challenges not only requires advanced Signal Processing and Machine Learning (ML) techniques but also emphasizes the critical need for interpretability, a requirement that remains a limitation of Deep Learning (DL) models~\cite{zhang2024artificial}.


\textcolor{black}{Several ML-based studies have highlighted the potential of these techniques in predicting MDR infections.  Models such as Logistic Regression (LR), XGBoost, Random Forest (RF), and Multilayer Perceptron (MLP) have been widely used to predict MDR bacterial colonization in hospitalized patients~\cite{ccaǧlayan2022data, li2024development, chen2022epidemiology, escudero2021use}. Typically, these approaches preprocess temporal information by aggregating and transforming it into static variables 
that summarize the patient's clinical history (e.g., most recent values, cumulative counts, or averages). Despite eliminating the temporal sequence of clinical events, such models have achieved promising predictive performance, with reported Receiver Operating Characteristic–Area Under the Curve (ROC-AUC) values ranging from 66.00\% for resistant urinary tract infections~\cite{rich2022development} to 77.50\% for hospitalized patients diagnosed with community-acquired pneumonia~\cite{rhodes2023machine}.}

\textcolor{black}{Recent advancements have analyzed Multivariate Time Series (MTS) data from patients' EHRs to predict single-pathogen MDR infections~\cite{hernandez2021antimicrobial}. Other studies have proposed Feature Selection (FS)-based ML frameworks to forecast infection outbreaks~\cite{jimenez2020feature} and applied temporal FS techniques to predict AMR, achieving an ROC-AUC of 72.06\%~\cite{escudero2020temporal}. In these approaches, temporal information is flattened to allow processing by classical ML models like Decision Tree (DT) and RF.}

\textcolor{black}{
Building on recent advances, DL methods have proven effective in modeling the temporal complexity of MDR. For example, \cite{hardan2024affordable} introduced a multimodal DL framework that used EHRs to predict Pseudomonas aeruginosa, achieving an ROC-AUC of 69.00\%. In parallel, several studies have investigated sequence-based DL models for analyzing MDR data derived based on clinical variables.  For instance, \cite{tharmakulasingam2023transamr} employed Transformer-based architectures for MDR prediction, achieving an ROC-AUC of 76.13\%. In contrast, \cite{martinez2023LSTM} integrated Dimensionality Reduction (DR) techniques with Long Short-Term Memory (LSTM) networks to predict resistance and susceptibility across six antibiotic classes, yielding an ROC-AUC of 67.70\%. However, these works generally lack interpretability mechanisms to support clinical decision-making and identify the key risk factors that contribute to the development of MDR.  
In response, recent research has integrated Explainable AI (XAI) techniques into DL frameworks to improve transparency in MDR prediction. However, methods such as SHapley Additive exPlanations (SHAP) often entail high computational costs and added complexity, which may limit their practical deployment in clinical settings~\cite{huang2024updates}. For instance, \cite{martinez2022interpretable} employed an LSTM architecture with SHAP that achieved an ROC-AUC of 66.73\%, enabling the identification of the most relevant variables to the classification task.}

\textcolor{black}{While ML and DL approaches have shown strong performance in predicting MDR, they often rely on complex architectures applied directly to raw clinical data. Although some efforts have introduced XAI to enhance interpretability, simpler and more transparent strategies—such as unsupervised clustering—remain largely underexplored for uncovering MDR-related risk factors. Moreover, most studies apply FS or Feature Engineering (FE) in isolation, with limited attention to patient-to-patient similarity derived from MTS data or the systematic use of DR techniques to mitigate overfitting and data correlation.}

\begin{figure}[t!]
    \centering
	\centering
	\includegraphics[width=1\columnwidth]{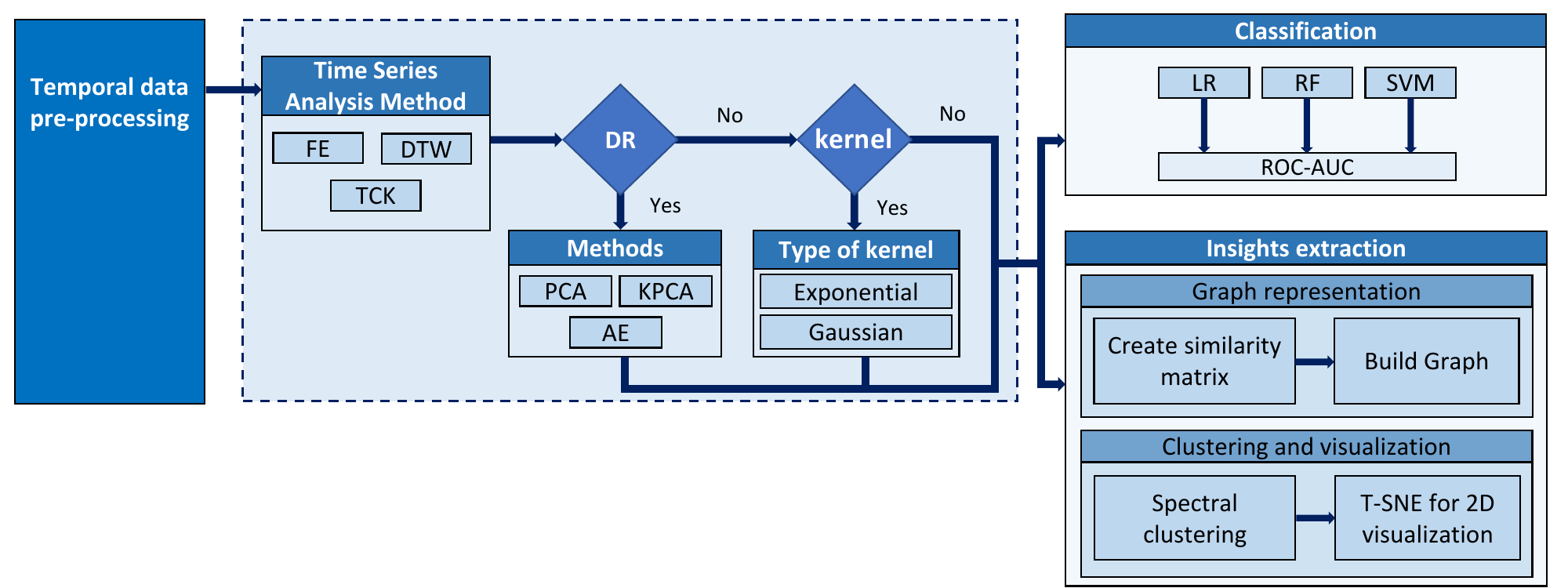}
	\caption{Architectural workflow that integrates time series analysis techniques with DR methods and kernel transformations, aimed at the classification of patients with MDR and the extraction of valuable insights, via graph representation, clustering, and visualization.
    }
	\label{fig:pipeline}
\end{figure}

\textcolor{black}{This work proposes an interpretable ML framework for MDR prediction and visualization.
Our key contribution is a patient-to-patient similarity representation based on MTS metrics, which quantify the similarity or distance between patients. Specifically, we model each patient as an MTS, where time-varying variables correspond to EHR data collected during the ICU stay. This similarity matrix is then transformed into a vector representation, enabling downstream analysis and integration into predictive models.} 
To that end, we select a random subset of $P$ patients and represent each patient by a vector of length $P$, with the $p$-th entry of the vector quantifying how similar the patient at hand is relative to the $p$-th patient in the preselected subset. Similarities or distances between patients are computed using time-aware functions (depicted in Figure~\ref{fig:pipeline} as the Time Series (TS) analysis method), which take two MTS matrices as input and output their level of similarity or distance. Three different functions are considered and tested, all of which account for the temporal nature of our data.

The first function is a simple \textcolor{black}{FE} approach, where each TS is replaced with a small number of time-dependent features (attributes computed using descriptive statistics), and patient similarities are then computed using the Euclidean inner product over the feature space~\cite{kocbek2019maximizing}. The second is Dynamic Time Warping (DTW)~\cite{berndt1994using}, which has been successfully applied in areas such as speech processing and is here extended here to deal with multiple series of different natures (numerical, categorical, and binary). The third and final method is Time Cluster Kernel (TCK), a technique that has demonstrated particular efficacy in clustering and classification tasks in healthcare settings~\cite{escudero2021use, mikalsen2021kernel}. Moreover, whenever convenient, we implement two additional preprocessing blocks: 1) a kernel block that can be used to transform distances into similarities, and 2) a DR block that yields smaller vector representations, preventing overfitting and naturally removing strong patient correlations present in the database. 

The resultant representations are then used for two different purposes: a) as input to a classifier whose output determines whether the patient is MDR or not, and b) as input to a graph-learning scheme whose output is used to visualize clinical patterns and cluster patients. Regarding MDR prediction, rather than employing large deep models, we focus on simple and interpretable architectures, including an LR, an RF classifier, and a Support Vector Machine (SVM) classifier. Intuitively, the weights of the regressor (or the branches in the \textcolor{black}{DT}) capture the relevance of each patient in the preselected dataset for predicting MDR. The underlying assumption in this approach is that the postulated MTS-based distances are powerful enough to capture the complexities of the dataset and the task at hand, early detection of MDR.

For the visual representation, we leverage the patient-to-patient similarity matrices to build a graph that identifies the most important connections between patients. Such a graph is then used to cluster patients, visualize them in a two-dimensional space (using techniques such as spectral clustering~\cite{von2007tutorial} and t-SNE~\cite{anowar2021conceptual}), 
and display information associated with patients as a graph signal. 

Overall, we propose an intuitive and relatively simple ML approach that offers both accurate predictions and meaningful insights. To validate our architecture, we deploy it on a database formed by EHRs from ICU patients at the University Hospital of Fuenlabrada (UHF) in Madrid, Spain. The records, collected over the last two decades, include more than 3,300 patients (18\% of whom acquired at least one MDR infection during their stay), 50 variables per patient, and data associated with their ICU stay (set to 7 days to facilitate comparisons). We test multiple versions of our architecture and provide visualizations to medical experts. The obtained results confirm prior existing medical findings, yield novel insights, and achieve MDR detection performance superior to state-of-the-art alternatives. While the \textcolor{black}{ROC-AUC} results are meaningful, we emphasize that an important contribution of our work is to highlight that in clinical applications with limited data of high quality, using interpretable patient-to-patient similarity representations—where the similarity metrics account for the heterogeneous and time-varying nature of the input data—has the potential to generate robust prediction results while facilitating representation and interpretability.

The primary contributions of this study, several of which can be extrapolated to other clinical problems involving MTS data, are:

\begin{itemize}
    \item \textcolor{black}{\textit{Development of an architecture for predicting MDR using MTS data.} We propose an innovative framework that overcomes the challenges of analyzing MTS data by combining advanced MTS techniques, DR methods, and classical classifiers (LR, RF, and SVMs). Our approach effectively captures critical temporal patterns in ICU EHRs, thereby enhancing the precision and reliability of MDR prediction.}
    
    \item \textcolor{black}{\textit{Patient-to-patient similarity representations and interpretability techniques.} Our framework leverages patient-to-patient similarity representations to simultaneously boost interpretability and maintain robust predictive performance.}
    
    \item \textcolor{black}{\textit{Graph-based methods for MDR pattern extraction.} By integrating DR, graph learning, and spectral clustering, we introduce novel graph-based visualizations that extract clinically actionable MDR patterns from patient data. To the best of our knowledge, this is the first study to employ such methods for both the extraction and characterization of MDR patterns.}
    
    \item \textcolor{black}{\textit{Empirical benchmarking.} We perform a comparison between our method and several state-of-the-art ML and DL models (including LSTM, Gated Recurrent Unit (GRU), and Transformer architectures), both standalone and integrated with preprocessing techniques. Our framework evaluates the models' ability to handle MTS data under varying assumptions—ranging from temporally structured to fully flattened representations—thus providing robust external baselines to validate the performance and reliability of our proposed method.}
\end{itemize}

The remainder of the article is organized as follows. Section~\ref{sec:Methods} presents the notation and methods used in this work. Subsequently, Section~\ref{sec:Database} describes the dataset used. The experiments and results are shown in Section~\ref{sec:DataAnalytic}. Finally, Section~\ref{sec:Disc} presents the main conclusions and associated discussions.

\section{Methods}\label{sec:Methods}

This section introduces the notation and provides a concise description of the data science methods comprising the architecture proposed in this work, as depicted in Figure~\ref{fig:pipeline}.

\subsection{Notation}

We represent our patient dataset as $\mathcal{D}=\{(\mathbf{X}_{i}, y_{i})\}_{i=1}^I $, where $I$ denotes the number of (labeled) patients. The $i$-th patient is represented by the feature matrix $\mathbf{X}_i\in \mathbb{R}^{F \times T}$, where $F$ is the number of features (variables) and $T$ the number of time slots. The $t$-th column of $\mathbf{X}_i$, denoted as $[\mathbf{X}_i]_{(:,t)}$, is a vector that contains the $F$ features of the $i$-th patient for the time slot $t$. Similarly, the $f$-th row of $\mathbf{X}_i$, denoted as $[\mathbf{X}_i]_{(f,:)}$, is the one-dimensional TS representing values of feature $f$ for patient $i$ over time. Patients who have at least one positive MDR culture during their ICU stay are assigned to the MDR class, while all other patients are classified as belonging to the non-MDR class. Given that we are dealing with a binary classification task, we have considered the label ``1'' to denote MDR patients and the label ``0'' to represent non-MDR patients. Finally, the label for the \mbox{$i$-th} patient is denoted as $y_i\in\{0,1\}$, while the soft label estimated (predicted) by the ML model is denoted as $\hat{y}_i\in[0,1]$.

\subsection{Time Series Analysis Methods}
\label{Methods_TSA}

This section briefly describes the three methods we employ to handle MTS, namely: FE, TCK, and DTW (see Figure~\ref{fig:pipeline}). All three methods discussed herein are unsupervised, meaning they do not utilize labels during data processing. Consequently, whenever we refer to the training set in this section, it is crucial to note that only the data $\{\mathbf{X}_i\}_{i=1}^I$ from $\mathcal{D}$ is considered, with the labels being disregarded.

\subsubsection{Feature Engineering}
\label{subsubsec:feature_engineering}

\textcolor{black}{FE is a common strategy for transforming complex data into fixed-length representations. In the context of TS and MTS a common approach is to summarize temporal dynamics through descriptive statistics~\cite{verdonck2021special}. This approach enables the use of conventional ML classifiers that do not require explicit temporal modeling while improving interpretability and reducing computational complexity. In our framework, each patient is represented by a matrix $\mathbf{X}_i \in \mathbb{R}^{F \times T}$. For each feature (i.e., row of $\mathbf{X}_i$), we compute five statistics: mean, median, mode, minimum, and maximum~\cite{kocbek2019maximizing}. This results in a transformed matrix of size $F \times 5$, effectively reducing the dimensionality from $F \times T$ while retaining key aggregate information. This fixed-size representation captures relevant temporal summaries, discards redundant fluctuations, and facilitates the use of interpretable ML models such as LR, RF, and SVM. Although it does not exploit fine-grained temporal patterns, FE offers a computationally efficient and clinically meaningful baseline within our architecture.}

\subsubsection{Time Cluster Kernel}

TCK is a method for evaluating the similarity between two-time sequences. TCK operates in an unsupervised manner, is capable of handling multiple features, and can automatically manage missing data—a significant limitation present in most clinical scenarios~\cite{mikalsen2018time, mikalsen2021time}. The fundamental concept behind TCK is to model the MTS as samples from a Gaussian Mixture Model (GMM)~\cite{mikalsen2018time}. The GMM parameters are learned, the posterior probabilities for each sample are computed, and a kernel matrix is then constructed. The number of mixture components, denoted as $N_C$, must be sufficiently large to capture both the local and global structures of the data, yet small enough to avoid overfitting. Based on the developers' recommendations and considering datasets similar to ours, for datasets with more than 100 samples, $N_C$ is set to 40. Each Gaussian component is defined by a mean vector, a diagonal covariance matrix (with the $f$-th element of the diagonal being the variance of the $f$-th feature), and a mixing coefficient. Therefore, the total number of parameters in the GMM model is $(2F + 1)N_C$~\cite{mikalsen2018time}. Once the GMM parameters are estimated using the expectation-maximization algorithm, each training sample (denoted as the $i$-th one) is mapped to an $N_C$-dimensional vector $\boldsymbol{\pi}_i \in [0,1]^{N_C}$, where the $n$-th entry represents the posterior probability of sample $i$ being generated by component $n$. The final step involves computing a kernel (similarity matrix) of size $I \times I$, where the $(i,j)$-th entry is obtained as the inner product of $\boldsymbol{\pi}_i$ and $\boldsymbol{\pi}_j$.

A notable feature of most TCK implementations is that instead of training a single GMM using the entire training set $\mathcal{D}_{train}:=\{\mathbf{X}_i\}_{i=1}^I$, random sampling techniques are employed to generate $K$ subsets from $\mathcal{D}_{train}$. Each subset is created by randomly selecting $I' \ll I$ observations, $F' < F$ variables, and $T' < T$ time instants from $\mathcal{D}_{train}$. Subsequently, the parameters of one GMM are estimated for each of the $K$ training subsets. For datasets similar to ours, the number of subsets $K$ is recommended to be set to 30. Each of the $K$ GMM models is then used to generate an $I \times I$ kernel matrix, and finally, the $K$ kernel matrices are aggregated (summed) to produce a single $I \times I$ output, where the $(i,j)$-th entry quantifies the similarity between MTS (patients) $i$ and $j$. It is important to note that the aggregated matrix is guaranteed to be positive semidefinite since each of the $K$ kernel matrices is itself positive semidefinite. Further details on the TCK method can be found in~\cite{mikalsen2018time}.

\subsubsection{Dynamic Time Warping}

DTW is a method for measuring dissimilarity between time sequences~\cite{muller2007dynamic}. Originally developed for speech signals, DTW was designed to manage misalignments between sequences~\cite{muller2007dynamic}. While most efforts focus on the one-dimensional case, its extension to multiple dimensions is less common~\cite{shokoohi2017generalizing}. 

To illustrate, let us assume that in our database, a patient's state is represented by a one-dimensional TS, $\mathbf{\underline{x}}_i = [{x}_i^{(1)}, {x}_i^{(2)},...,{x}_i^{(T)}] \in \mathbb{R}^{1 \times T}$~\cite{seto2015multivariate}. We first explain how to compute DTW for one-dimensional series and then extend the method to the multidimensional case. Consider two one-dimensional TS $\mathbf{\underline{x}}_1$ and $\mathbf{\underline{x}}_2$ of length $T$, corresponding to patient $i=1$ and patient $i=2$. To compute $DTW(\mathbf{\underline{x}}_1, \mathbf{\underline{x}}_2)$, we construct a cumulative distance matrix $\mathbf{M} \in \mathbb{R}^{(T+1) \times (T+1)}$ using the following initialization and recursive procedure:
\begin{align}
[\mathbf{M}]_{1,1}&=0,~[\mathbf{M}]_{1,t+1}=\infty,~[\mathbf{M}]_{t+1,1}=\infty,~\forall t, \label{eq:Matrix_DTW_step1}\\
[\mathbf{M}]_{t,t'} &= \delta([\mathbf{\underline{x}}_1]_{t}, [\mathbf{\underline{x}}_2]_{t'}) + \min\{[\mathbf{M}]_{t-1,t'-1}, [\mathbf{M}]_{t-1,t'},  [\mathbf{M}]_{t,t'-1}\}, \label{eq:Matrix_DTW_step2}
\end{align}
where $\delta: \mathbb{R}\times \mathbb{R}\rightarrow [0,\infty)$ is any scalar distance function. After completing $\mathbf{M}$ column by column (or row by row), the DTW distance is obtained as $DTW(\mathbf{\underline{x}}_1, \mathbf{\underline{x}}_2) = [\mathbf{M}]_{T+1,T+1}$~\cite{seto2015multivariate}.

Next, we define the DTW distance between two MTS $\mathbf{X}_1$ and $\mathbf{X}_2$. There are two approaches in the literature for carrying out this task\footnote{A library, developed by the authors of this work, is available for the efficient implementation of Dynamic Time Warping and its application to MTS (in any field of application beyond the clinical one)~\cite{escudero2023dtwparallel}. The library represents an improvement in computational time, interpretation, and understanding of the method compared to existing ones, allowing for a more didactic use. The code is available on GitHub: {\url{https://github.com/oscarescuderoarnanz/dtwParallel}}.}: “Dependent DTW” denoted as $DTW_D: \mathbb{R}^{F \times T} \times \mathbb{R}^{F \times T} \rightarrow [0,\infty)$ and “Independent DTW,” denoted as $DTW_I: \mathbb{R}^{F \times T} \times \mathbb{R}^{F \times T} \rightarrow [0,\infty)$~\cite{shokoohi2017generalizing}. The idea behind $DTW_D$ is to modify the step \eqref{eq:Matrix_DTW_step2} when computing the matrix $\mathbf{M}$. Specifically, instead of assuming that the arguments of $\delta$ are scalars, the updated distance function $\delta: \mathbb{R}^F \times \mathbb{R}^F \rightarrow [0,\infty)$ is used. With this in mind, if one is interested in finding $DTW_D(\mathbf{X}_1,\mathbf{X}_2)$, the $DTW_D$ distance between the MTS representing patients $i=1$ and $i=2$, one can run \eqref{eq:Matrix_DTW_step1}-\eqref{eq:Matrix_DTW_step2} with $\delta([\mathbf{X}_1]_{(:,t)}, [\mathbf{X}_2]_{(:,t')})$ (choosing, e.g., the Euclidean norm in $\mathbb{R}^F$ as the function $\delta$), and then set $DTW_D(\mathbf{X}_1,\mathbf{X}_2)$ to $[\mathbf{M}]_{T+1,T+1}$. 

On the other hand, the idea behind $DTW_I$ is to consider the distance between two $F$-dimensional MTS as the sum of $F$ DTW distances, each computed independently for each of the $F$ features. Formally, with $[\mathbf{X}_i]_{(f,:)}$ denoting the $f$-th row of the matrix $\mathbf{X}_i$, the distance between the MTS representing patients $i=1$ and $i=2$ is obtained as $DTW_I(\mathbf{X}_1, \mathbf{X}_2)= \sum_{f=1}^{F}DTW([\mathbf{{X}}_1]_{(f,:)}, [\mathbf{{X}}_2]_{(f,:)})$.

Given a training dataset with $I$ patients, we compute (either the dependent or independent version of) $DTW(\mathbf{X}_i,\mathbf{X}_j)$ for every pair of patients $(i,j)$. The values are then arranged into an $I\times I$ symmetric distance matrix whose $(i,j)$-th entry denotes the (either dependent or independent) DTW distance between patients $i$ and $j$.

\subsection{Dimensionality Reduction via Feature Transformation}\label{sec:subsec_DR}

This section details the methods employed to reduce the dimensionality of the data (refer to Figure~\ref{fig:pipeline}), specifically Principal Component Analysis (PCA), Kernel Principal Component Analysis (KPCA), and Autoencoder(AE).

\subsubsection{Principal Component Analysis}

PCA is an unsupervised DR method that linearly projects data from a higher-dimensional space onto a lower-dimensional subspace, resulting in new features known as Principal Components (PCs)~\cite{anowar2021conceptual}. PCA achieves this by utilizing the leading eigenvectors of the covariance matrix of the data as the projection matrix. This method is optimal under several criteria, one of which is the minimization of the quadratic error between the original data and its low-dimensional representation~\cite{van2009dimensionality}. 

\subsubsection{Kernel Principal Component Analysis}

KPCA is a non-linear extension of PCA. Unlike PCA, KPCA applies a non-linear transformation to map the data into a higher-dimensional space, where the PCs are then computed. This procedure involves calculating the PCs from the \emph{kernel} matrix instead of directly from the covariance matrix~\cite{van2009dimensionality}. In this work, we experiment with different types of \emph{kernels} and their corresponding hyperparameters to identify the optimal configuration. The optimal kernel type and its associated hyperparameters are those that minimize the quadratic error between the original dataset and the reconstructed data in the reduced-dimensional space for the specific dataset under consideration.

\subsubsection{Autoencoders}

AE is an unsupervised neural network architecture designed with two main objectives: (i) to encode the input data into a compressed and meaningful latent representation, and (ii) to decode this compressed representation back into the original data space, ensuring that the reconstructed output closely resembles the original input~\cite{bank2020autoencoders}. The encoder compresses the input data into a lower-dimensional latent space, while the decoder reverses this process, mapping the latent space back to the higher-dimensional input space. The AE is trained through an iterative optimization process aimed at minimizing the training cost function, specifically the Mean Squared Error (MSE) between the input of the encoder and the output of the decoder. During each iteration, a subset of the data is used as input, and the error between the encoded-decoded output and the original data is back-propagated through the network to update the weights. AEs are primarily employed as a DR tool, providing a low-dimensional representation of the original data by mapping it onto the latent space~\cite{hinton2006reducing}. 

Additionally, Denoising Autoencoders (DAE) serve as a regularization variant of traditional AEs, enhancing robustness to errors by introducing noise during training~\cite{bank2020autoencoders}. In this study, we utilize DAE by adding white Gaussian noise with zero mean and unit variance to the inputs, thereby improving the network's capacity for error correction.

\subsection{Machine Learning Classifiers}

In this subsection, we discuss three ML classifiers that are fundamental to our approach: LR, RF, and SVM.

\subsubsection{Logistic Regression}

LR is a ML technique that models binary labels through a logistic function. The coefficients of the model are determined by minimizing the Binary Cross-Entropy cost function, which quantifies the disparity between predicted and actual labels. In the formal LR framework, each sample is represented by a feature vector $\mathbf{x}_i$ of $F$ dimensions. The LR model requires $F+1$ parameters, with $F$ of them corresponding to the weights in the vector $\mathbf{w}\in\mathbb{R}^{F}$. To calculate the LR weights, an optimization procedure is initiated, which includes a regularization term to prevent overfitting. This optimization task necessitates fine-tuning the parameters to strike a balance between regularization and training error, with the hyperparameter linked to the regularizer playing a crucial role. Typically, gradient descent is utilized to solve this convex optimization problem. It is important to note that when dealing with non-linearly separable data, the model may encounter slow convergence.

\subsubsection{Random Forest}

RF is a supervised ML approach commonly used for non-linear classification and regression tasks~\cite{fraiwan2012automated}. It consists of an ensemble of $K$ classification trees, each independently trained on a randomly generated subset of the training data using the bagging technique~\cite{dai2018using}. Each tree provides a prediction for the class label, and these individual predictions are aggregated, typically through majority voting, to determine the final output. The bagging technique mitigates overfitting by generating $K$ subsets from the training set $\mathcal{D}_{train}$, where each subset is formed by randomly selecting $I'\ll I$ observations and $F'<F$ features. This approach enhances the model’s robustness to fluctuations in the input data, improving stability and accuracy~\cite{ExtraTreeClassifier}. Ultimately, the outputs from the $K$ trees are combined to produce the final prediction, making RF particularly effective for complex datasets.

\subsubsection{Support Vector Machine}

SVM are supervised learning algorithms used for classification and regression tasks~\cite{suthaharan2016support}. They define a hyperplane as a decision boundary that maximizes the margin—the distance between the hyperplane and the nearest data points. This approach involves input vectors (\(\mathbf{x}_i\)), weights (\(\mathbf{w}\)), and a bias term (\(b\)), with the hyperplane in input space given by \(h(\mathbf{x}_i\;|\mathbf{w},b) = \mathbf{w}^{\top}\mathbf{x}_i + b\). For non-linearly separable data, SVMs use a ``kernel'' method to compute scalar products in a transformed space, enabling a linear boundary without explicit transformation. Common kernels include radial basis and sigmoid~\cite{suthaharan2016support}. Regardless of the chosen kernel, the optimization problem is convex and solvable via gradient descent. For more details on kernel methods, see~\cite{widodo2007support}. The \(\nu\)-SVM variant introduces \(\nu \in [0, 1]\) to control the fraction of training errors and support vectors directly, providing greater flexibility than the standard penalty parameter \(C\)~\cite{suthaharan2016support}. This allows \(\nu\)-SVM to achieve a controlled trade-off between margin and classification errors, proving especially effective in noisy or complex data scenarios.

\subsection{Tools to Visualize and Extract Knowledge from Data}
\label{visualization}

Gaining intuition from high-dimensional data presents significant challenges. This section outlines two powerful tools—t-SNE and spectral clustering—that aid in visualizing complex datasets and extracting meaningful insights (see Figure~\ref{fig:pipeline}). First, we employ graph-based spectral clustering to segment patients into distinct groups based on their similarities. Subsequently, we utilize t-SNE, a non-linear DR technique, to map the patient data into a two-dimensional space, enabling straightforward visualization. This approach not only facilitates the representation of different patient groups with distinct colors or markers but also aids in partitioning the 2D space into clearly defined regions. Visualization tools such as these are crucial in data science, particularly in clinical applications where extracting actionable knowledge is essential.

\subsubsection{t-SNE}\label{subsec-tsnemethods}

t-SNE is a non-linear, unsupervised method designed to map high-dimensional data into a lower-dimensional space—typically two or three dimensions—while preserving the significant structure of the data~\cite{anowar2021conceptual}. 
This method is widely used for data exploration and visualization~\cite{anowar2021conceptual}. In our study, t-SNE is applied to transform the high-dimensional patient-similarity feature space into a 2D Euclidean space (see Figure~\ref{fig:pipeline}). Formally, for a patient represented by the high-dimensional vector $\mathbf{x}_i\in\mathbb{R}^{F}$, t-SNE first models the data as multivariate Gaussian distributions ($G_1,...,G_I$), where each $i$-th Gaussian is centered at $\mathbf{x}_i$ with a common standard deviation $\sigma$. The similarity between patients is then calculated using conditional probabilities $P_{(\mathbf{x}_j|\mathbf{x}_i)}= \frac{G_i (\mathbf{x}_i, \mathbf{x}_j)}{\sum_{i'\not=i} G_i (\mathbf{x}_i', \mathbf{x}_i)}$, where $G_i(\mathbf{x}_j, \mathbf{x}_i) =  e^{\frac{-\| \mathbf{x}_j - \mathbf{x}_i \|^2}{2\sigma^2}}$. Next, in the reduced 2D space, each patient is represented by a two-dimensional vector $\mathbf{z}_i$, and similarities are modeled using a t-Student distribution. The divergence between the high-dimensional and low-dimensional similarities is measured using Kullback-Leibler (KL) divergence, which is minimized via gradient descent, yielding a 2D representation for each patient~\cite{anowar2021conceptual}. This results in a set of $I$ 2D points that can be easily visualized, as shown in Section~\ref{sec:experiments_visualization_tools}.

\subsubsection{Spectral Clustering}\label{sc}

Spectral clustering, rooted in algebraic graph theory, aims to partition a dataset into subsets while minimizing dissimilarity between connected nodes~\cite{von2007tutorial}. The process reframes the clustering task as a graph partitioning problem. Here, each patient is represented as a node in a graph, with edges connecting nodes that exhibit similarity. The objective is to minimize the number of connections, or ``cuts'', between nodes that belong to different clusters. This relaxed combinatorial problem is optimally solved by analyzing the eigenvectors of the Laplacian matrix associated with the graph~\cite{von2007tutorial}. Spectral clustering involves three primary stages: preprocessing, spectral representation, and clustering. In preprocessing, a graph is constructed with patients as nodes, and connections are established using a $K$-nearest neighbors approach based on the inverse of the Euclidean distance between patient representations. In the spectral representation stage, the Laplacian matrix $\mathbf{L}$ of the $K$-nearest neighbors graph is computed, followed by its eigendecomposition. Nodes are then assigned spectral features corresponding to the $C+1$ smallest eigenvalues of $\mathbf{L}$, where $C$ is the desired number of clusters. Finally, the patients are clustered based on this spectral representation~\cite{jia2014latest}.

Selecting the optimal number of clusters $C$ is a complex task. To address this, we employ Cluster Validity Indexes (CVIs) to identify the most appropriate value of $C$. These CVIs assess both intra-cluster similarity and inter-cluster separation, combining them into a quality indicator. In this work, we utilize the Silhouette index, which measures similarity based on the distance between all observations within the same cluster and separation based on the distance to the nearest neighboring cluster~\cite{arbelaitz2013extensive}. We also use the Davis-Bouldin index, which estimates similarity by the distance of points within a cluster to its centroid and separation by the distance between centroids~\cite{arbelaitz2013extensive}.

\section{Results}
\label{sec:DataAnalytic}

This section introduces the dataset, details the experimental setup, and evaluates the proposed MDR prediction approach, which integrates MTS analysis, DR, and kernel-based transformations. \textcolor{black}{We also compare our approach with state-of-the-art ML and DL models on the same dataset}. Additionally, we employ nonlinear embeddings and graph-based spectral clustering for data visualization, uncovering key MTS patterns relevant to MDR prediction. All experiments and results are publicly available at{\small~\url{https://github.com/oscarescuderoarnanz/DM4MTS}}, ensuring transparency and reproducibility.

\subsection{Database and Data Preparation}
\label{sec:Database}

\textcolor{black}{This study leverages a dataset of 3,310 ICU patients admitted between 2004 and 2020. Data were collected under a research agreement established in July 2020 between King Juan Carlos University and the UHF. The primary aim of this agreement was to enable translational research on ICU data, including medication records, microbiological cultures, and Mechanical Ventilation (MV) events. Ethical approval was granted by the UHF Research Ethics Committee (ref: 24/22, EC2091).}

\textcolor{black}{To ensure consistent MDR case labeling, we adopted the widely accepted definition proposed by the European Centre for Disease Prevention and Control and the U.S. Centers for Disease Control and Prevention~\cite{magiorakos2012multidrug}, which defines MDR as resistance to at least one agent in three or more antimicrobial categories. This criterion was retrospectively applied across the full study period, and all microbiological records were reviewed and annotated accordingly. The dataset includes 597 patients who developed MDR and 2,713 who did not.}

\textcolor{black}{Each patient is described by both static and temporal variables. Static data include demographic information, severity indices, and admission/discharge unit details, while temporal data capture events during the ICU stay, such as MV administration and antibiotic usage—key indicators in the development of MDR. Notably, a protocol change in microbiological screening was introduced in June 2013. Prior to this, screening was limited to high-risk patients upon ICU admission based on a predefined checklist and involving nasal, pharyngeal, rectal, and axillary swabs. After the change, systematic screening was extended to all ICU admissions, regardless of risk profile, using nasal, pharyngeal, and rectal swabs, with wound cultures added when clinically relevant. Weekly surveillance cultures were performed every Tuesday for all ICU patients, both before and after the protocol shift. To mitigate potential biases arising from changes in detection sensitivity, we ensured a proportional representation of patients from both periods during model training.}

\begin{figure}[h!]
    \centering
	\includegraphics[width=0.7\columnwidth]{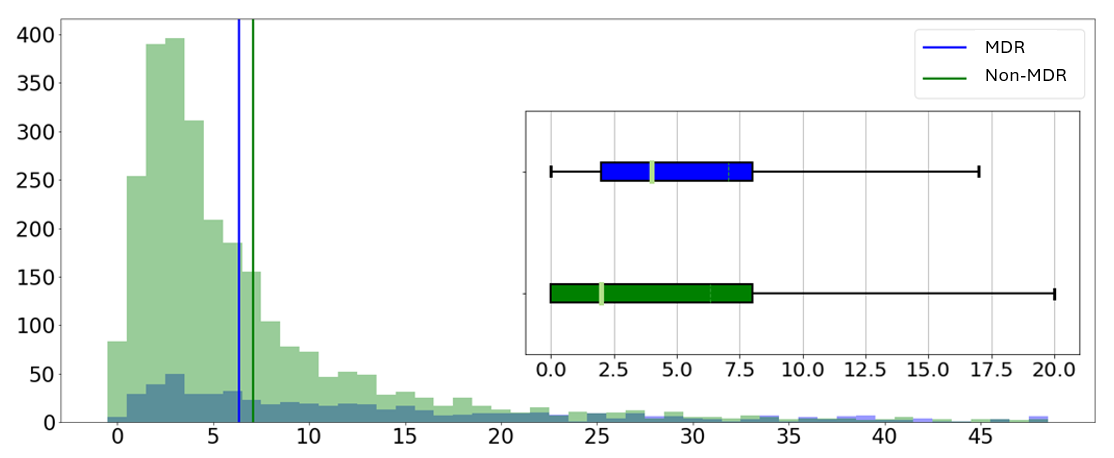}
	\caption{Histogram and boxplots of the time elapsed (in days) from ICU admission to ICU discharge (delimited by 50 days). The first vertical line (blue) corresponds to MDR patients and indicates the average time from ICU admission to the first MDR acquisition. The second vertical line (green) represents the average stay length for non-MDR patients.}
	\label{fig:lenwindow}
\end{figure}

\textcolor{black}{Given our focus on ICU-acquired MDR, we excluded pre-ICU information such as ward-level or outpatient data. This decision, guided by ethical considerations and clinical relevance, emphasizes transmission and treatment patterns specific to the ICU, where the risk of nosocomial MDR acquisition is highest. Although external data could offer additional context, the ICU dataset already includes rich, high-frequency clinical and microbiological information that sufficiently supports predictive modeling. Finally, to address variability in time sampling—common in real-world EHRs~\cite{xiao2018opportunities}—we standardized all temporal sequences to a 24-hour interval, in line with standard practice and previous work~\cite{martinez2022interpretable}. We fixed the series length to \(T = 7\) days, based on the typical ICU stay duration for non-MDR patients and the median time to first MDR detection in positive cases (see Figure~\ref{fig:lenwindow}), thus ensuring temporal alignment, reducing missingness, and facilitating model integration.}

To create the TS characterizing each patient, time is indexed from $t_0$ to $t_6$ (i.e., $T=7$ days). For MDR patients, the last time slot corresponds to the day of the positive culture, with the culture being identified as MDR 24 to 48 hours later. For non-MDR patients, $t_0$ corresponds to the time of ICU admission. The TS is padded with zero values if the ICU stay is shorter than 7 days for non-MDR patients or if the event occurs within the first 7 days for MDR patients. In this context, the event is defined as the first positive culture result.

Each patient is characterized by an array of TS, all of which have the same length. The first TS corresponds to MV, and it consists of a sequence of binary values that indicate, for each time slot, whether a specific patient has received MV or not. Then, we use TS associated with different antibiotic families, namely: Aminoglycosides (AMG), Antifungals (ATF), Carbapenems (CAR), 1st generation Cephalosporins (CF1), 2nd generation Cephalosporins (CF2), 3rd generation Cephalosporins (CF3), 4th generation Cephalosporins (CF4), Glycyclines (GCC), Glycopeptides (GLI), Lincosamides (LIN), Lipopeptides (LIP), Macrolides (MAC), Monobactams (MON), Nitroimidazolics (NTI), Miscellaneous (OTR), Oxazolidinones (OXA), Broad-Spectrum Penicillins (PAP), Penicillins (PEN), Polypeptides (POL), Quinolones (QUI), Sulfamides (SUL), Tetracyclines (TTC), and unclassified antibiotics (Others). The ``Others'' family of antibiotics identifies any other antibiotic not considered in the previous families. Thus, for the $i$-th patient, and similarly to the MV variable, the $f$-th feature linked to antibiotics is a binary sequence identifying whether that particular antibiotic has been taken during each $24$-hour period (time slot). This entails 23 variables. 
In addition to these, we generated an additional set of 25 numerical temporal features for each patient, referred to as ``environmental variables''. These features act as indicators of the health status of the co-patients within the ICU. Co-patients are those ICU patients staying in the ICU during the same time slot as the $i$-th patient, excluding the patient under study. Environmental variables are linked to: i) the number of total ICU co-patients, ii) the number of MDR co-patients, and iii) the number of co-patients taking each of the 23 families of antibiotics. The incorporation of these environmental ICU variables, a key aspect of our approach, is well-supported by clinical insights and has been demonstrated to be advantageous in previous studies focused on MDR classification~\cite{martinez2022interpretable}.

\subsection{Experimental Setup}
\label{sec:ExperimentalSetup}

The original dataset $\mathcal{D}$ was randomly partitioned into two non-overlapping subsets, with 70\% of the samples allocated to the training set $\mathcal{D}_{train}$ and the remaining 30\% to the test set $\mathcal{D}_{test}$. To ensure robustness and generalizability, each experiment was repeated five times, with different random partitions of the training and test subsets. The training set was utilized for model development, while performance evaluation was conducted on the test set. As discussed in Section~\ref{sec:Database}, the dataset exhibits class imbalance, a common challenge in binary classification tasks. This imbalance can bias models towards the majority class, potentially compromising their generalization performance~\cite{wang2016training}. To address this issue, we employed an undersampling strategy to balance the classes within $\mathcal{D}_{train}$~\cite{he2009learning}. Additionally, a 5-fold cross-validation approach was employed during training to optimize the hyperparameters of the DR methods and ML models, as well as to prevent overfitting.

Min-max normalization was applied to $\mathcal{D}_{train}$, with the same scaling parameters subsequently applied to $\mathcal{D}_{test}$ to ensure consistency across the datasets. The cross-validation procedure was meticulously designed to select hyperparameters that best align with the data-model relationship, with a primary focus on minimizing overfitting. The ROC-AUC metric was maximized during hyperparameter tuning to achieve optimal model performance.

In terms of performance metrics, the DR methods, KPCA, and AE were optimized to minimize the MSE between the original space and the latent space. Conversely, for the ML models, the objective was to maximize the ROC-AUC, a metric that evaluates the overall performance of a binary classifier and provides insights into the trade-off between specificity and sensitivity~\cite{bradley1997use}. Specifically, for KPCA, both the type of kernel (exponential or radial basis) and the dimensionality of the reduced space were fine-tuned. In the case of AEs, hyperparameters such as the number of neurons (ranging from 250 to 450 in steps of 25), the learning rate $\{1\mathrm{e}{-4}, 1\mathrm{e}{-3}, 1\mathrm{e}{-2}, 1\mathrm{e}{-1}\}$, and the dropout rate \{0, 0.05, 0.1\} were optimized. For DAE, the noise level was also fine-tuned \{0.01, 0.025, 0.05\}. These models were designed with adaptive learning rates and early stopping to further regularize the training process. In contrast, PCA retained only as many PCs as necessary to explain 99\% of the variance. Given the mixed nature of the features in our dataset, the distance function $\delta$ in the DTW implementation [cf. \eqref{eq:Matrix_DTW_step2}] was set to the Gower distance~\cite{tuerhong2014gower}.

The hyperparameters for the ML models were optimized as follows: for LR, the weight regularization coefficient was explored across a range of values 
\{1e{-7}, 1e{-6}, 1e{-5}, 1e{-4}, 1e{-3}, 5e{-3}, 1e{-2}, 5e{-2}, 1e{-1}, 5e{-1}, 7.5e{-1}, 1, 3, 5, 8, 10, 12, 15\}; for RF, the maximum tree depth (from 10 to 42 in steps of 4), the minimum number of samples per leaf \{2, 4, 9, 13, 17, 22, 43\}, and the number of estimators \{30, 50, 100, 200, 400, 600\} were optimized; and for $\nu$-Support Vector Machine ($\nu$-SVM), both the percentage of support vectors $\nu$ \{1e{-4}, 1e{-3}, 5e{-3}, 1e{-2}, 5e{-2}, 1e{-1}, 5e{-1}, 7.5e{-1}, 9e{-1}\} and the gamma parameter \{1e{-8}, 1e{-7}, 1e{-6}, 1e{-5}, 1e{-4}, 1e{-3}, 1e{-2}, 5e{-2}, 1e{-1}, 1\} were fine-tuned. The comprehensive availability of hyperparameters, combined with the detailed description of data normalization, ensures the reproducibility and transparency of the proposed architecture. Furthermore, the open-source availability of the code reinforces these goals. Additional details on the experiments can be found in the supplementary material.

\subsection{MDR Prediction Using Temporal Features}
\label{sec:MDRprediction}

\textcolor{black}{In this section, we provide a comprehensive evaluation of our framework for MDR prediction using MTS data (see Section~\ref{sec:Database}). Performance is assessed via the ROC-AUC metric, and rather than reporting a single ROC-AUC value for each pipeline, we offer a robust statistical summary—including the median, 25th and 75th percentiles, as well as minimum and maximum values. This distribution-aware approach, illustrated through boxplots, provides deeper insights into the variability and stability of each pipeline's predictive capabilities across multiple splits of the dataset.}

\begin{figure}[!h]
    \centering
    	\begin{subfigure}[]
    		\centering
    		\includegraphics[width=13cm,height=3cm]{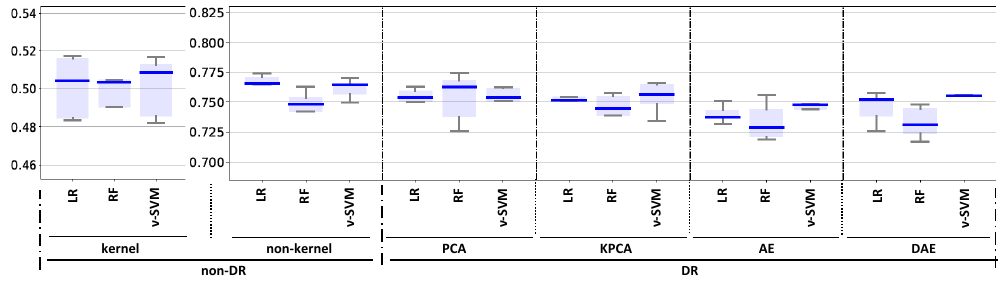}
    	\end{subfigure}
    	
    	\begin{subfigure}[]
    		\centering
    		\includegraphics[width=13cm,height=3cm]{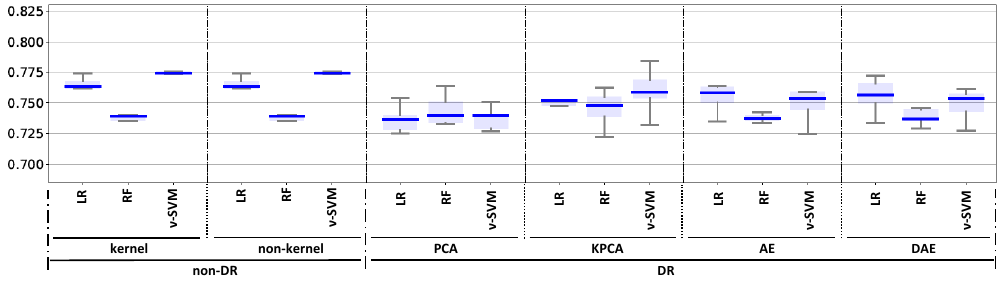}
    	\end{subfigure}
    	
    	\begin{subfigure}[]
    		\centering
    		\includegraphics[width=13cm,height=3cm]{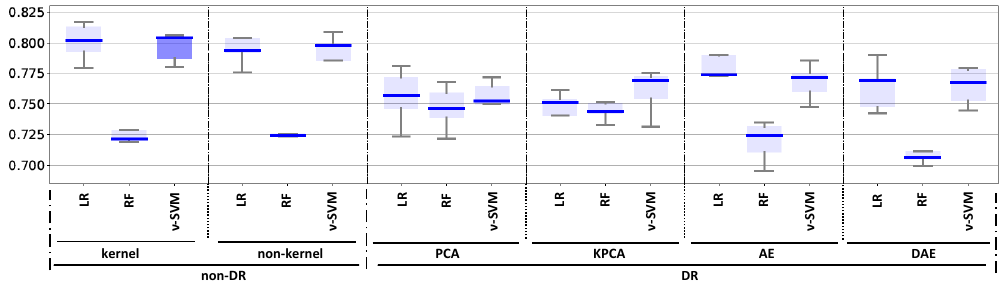}
    	\end{subfigure}
    	
    	\begin{subfigure}[]
    		\centering
    		\includegraphics[width=13cm,height=3cm]{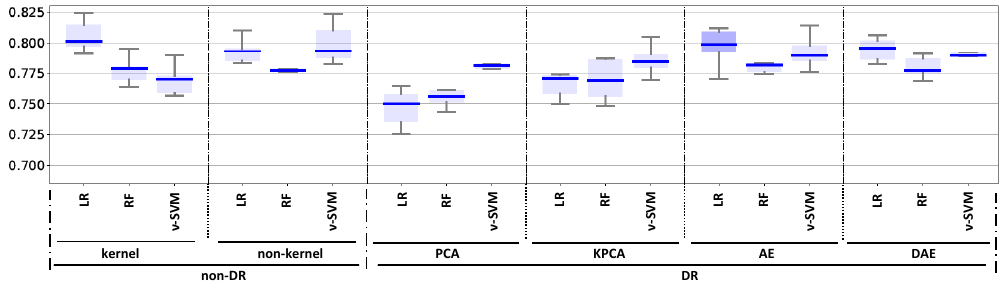}
    	\end{subfigure}
    	
    \caption{ROC-AUC values for the classification models (LR, RF, and $\nu$-SVM) when considering non-DR (original space and kernel transformations) and DR methods (PCA, KPCA, AE, and DAE) for: (a) FE; (b): TCK; (c) DTW$_D$; (d) DTW$_I$. Box-plots with the best results in terms of median ROC-AUC with and without DR have been highlighted in dark blue.}
	\label{fig:resultsPrediction}
\end{figure}

The outcomes of this analysis are presented across 24 panels, as depicted in Figure~\ref{fig:resultsPrediction}, organized into a $4\times 6$ layout. Each row within this arrangement represents a different approach to handling MTS data, denoted as FE (row (a)), TCK (row (b)), DTW$_D$ (row (c)), and DTW$_I$ (row (d)). The first two columns correspond to pipelines that do not incorporate DR, while columns 3 to 6 encompass four distinct DR methods: PCA, KPCA, AE, and DAE, respectively. Furthermore, for each of the 24 panels, we report ROC-AUC values associated with three different classification models: LR, RF, and $\nu$-SVM. To facilitate fair comparisons, we maintain a uniform dynamic range (axis) for all panels in the figures, except for the results of FE with a kernel transformation, which are displayed in the leftmost panel of Figure~\ref{fig:resultsPrediction} (a) due to considerably lower ROC-AUC values.

Our analysis begins by discussing the pipelines that do not incorporate DR (columns 1 and 2). For ease of presentation, when contrasting various approaches, we primarily emphasize the median ROC-AUC, incorporating other statistical metrics when deemed relevant. The principal observation for non-DR pipelines is that those employing DTW (refer to Figure~\ref{fig:resultsPrediction} (c) and (d)) outperform FE and TCK approaches (refer to Figure~\ref{fig:resultsPrediction} (a) and (b)). The least favorable results are associated with the FE-kernel combination (median ROC-AUC just above 50\%), while the most promising performance is observed for the DTW$_D$, exponential kernel, and $\nu$-SVM combination, yielding an ROC-AUC value of approximately 81\% (refer to Figure~\ref{fig:resultsPrediction} (c)). Remarkably, this outcome exceeds the previous best performance achieved on the same dataset~\cite{martinez2022interpretable}.

Focusing on pipelines implementing DR (columns 3 to 6), we discern that PCA and KPCA (columns 3 and 4) yield comparable outcomes, with a slight preference for the former. Similarly, AE and DAE have no significant performance distinctions (columns 5 and 6). However, both AE and DAE outperform their linear and non-linear counterparts, PCA and KPCA. When comparing the different methods for processing MTS data (rows 1-4), the results confirm that, similar to non-DR pipelines, superior performance is achieved when using DTW$_D$ and DTW$_I$. The most favorable combination in this context comprises DTW$_I$, AE, and LR, boasting a median ROC-AUC value of 80\%. Interestingly, while the poorest-performing DR method yields a median performance of 70\% (significantly surpassing the 50\% mark for the non-DR case), the best median performance for DR pipelines (80\%) is marginally lower than the pinnacle performance for non-DR pipelines (81\%).

Based on the aforementioned findings and insights provided in Figure~\ref{fig:resultsPrediction}, we derive several pertinent conclusions regarding the analysis of MTS data: (i) DTW emerges as the optimal method for handling MTS data; (ii) when TCK and non-DR are considered, the interquartile ranges, a measure of the spread of the data obtained as the difference between the 75th and 25th percentiles of the boxplots, are smaller (see Fig~\ref{fig:resultsPrediction} (b)); and (iii) the incorporation of DR techniques does not yield substantial benefits. 

The previous findings highlight the importance of using MTS-based metrics for evaluating the similarities between patients. This is not only relevant from a classification perspective but also due to their ease of interpretation. On the other hand, it is important to recognize that a patient-based similarity representation yields high-dimensional vectors that may not be easy to visualize. The following section offers various alternatives to address this specific challenge.

\subsubsection{\textcolor{black}{Validation and Benchmarking Experiments}}
\label{sec:new_exp}
\begin{figure}[h!]
    \centering
	\includegraphics[width=\columnwidth]{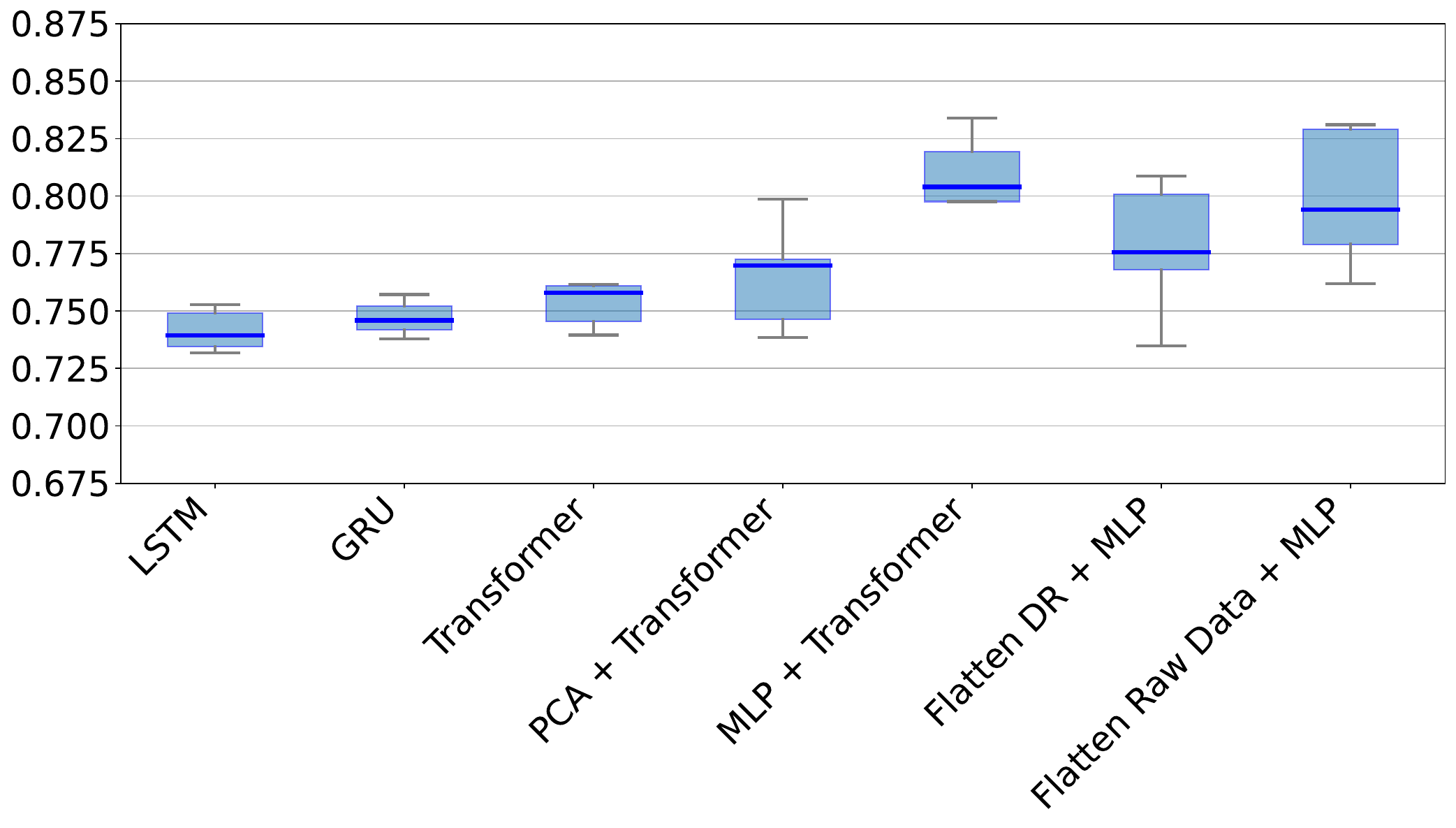}
    \caption{\textcolor{black}{Box plots of the ROC-AUC scores for each evaluated model (x-axis), with the median values highlighted in blue. The boxes represent the interquartile range (y-axis). The models are grouped as follows: LSTM, GRU, and Transformers correspond to \textit{Raw MTS to DL models}; PCA + Transformer represents \textit{Feature PCA-DR + Transformer}; MLP + Transformer denotes \textit{Feature MLP-DR + Transformer}; Flatten DR + MLP refers to \textit{Hybrid row-column DR + Flattening and MLP}; and Flatten Raw Data + MLP corresponds to \textit{Raw data + Flattening and MLP}.}}

	\label{fig:new_results}
\end{figure}

\textcolor{black}{
We benchmark state-of-the-art DL models on our dataset to establish a baseline. We design a framework to assess how different architectures impact MTS modeling, from temporally-aware to agnostic approaches. The pipeline starts with raw MTS inputs, preserving full temporal dynamics. Next, a feature-wise DR step compresses the input while retaining temporal structure. We then evaluate models that ignore temporal order but preserve spatial relations, followed by a hybrid setup combining spatial and temporal modeling. Finally, we test a fully agnostic representation that flattens both dimensions. These configurations, used as external benchmarks, enable a thorough assessment of our framework’s performance (see Figure~\ref{fig:new_results}).
}

\begin{enumerate}[leftmargin=*]

    \item \textcolor{black}{\textit{Raw MTS to DL models:} In this baseline configuration, the MTS $\mathbf{X}_i \in \mathbb{R}^{F \times T}$ is used in its raw form, preserving the complete temporal structure. We evaluate three widely adopted sequential architectures—LSTM, GRU, and Transformer—chosen for their proven capacity to capture long-term dependencies and dynamic temporal patterns. The best ROC-AUC scores achieved were 74.14\% (LSTM), 74.69\% (GRU), and 75.30\% (Transformer).}
    
    \item \textcolor{black}{\textit{Feature PCA-DR + Transformer:} To reduce feature redundancy and enhance model generalization, we apply PCA independently at each time step, so that each column $[\mathbf{X}_i]_{(:,t)}$ is projected onto a lower-dimensional subspace of dimension $F^*$ with $F^*\ll F$. Each column gives rise to the latent representation $\mathbf{z}_{i,t}\in\mathbb{R}^{F^*}$. The $T$ signals are then collected into the ${F^* \times T}$ matrix $\mathbf{Z}_i$. This dimensionality reduction retains the most informative components while preserving the temporal structure of the data. The resulting sequence is then processed using a Transformer, selected based on its superior performance in our ``Raw MTS to DL models'' benchmark. This combination is particularly beneficial in high-dimensional settings, where mitigating noise and overfitting is critical to model robustness. Under this configuration, we achieved an ROC-AUC of 76.51\%.}
    
    \item \textcolor{black}{\textit{Feature MLP-DR + Transformer:} We replicate the processing in the previous section, but encoding each $[\mathbf{X}_i]_{(:,t)}$ separately using a two-layer feedforward network, giving rise to the embedding $\mathbf{z}_{i,t} \in \mathbb{R}^{F^*}$. The $T$ embeddings are concatenated into the  ${F^* \times T}$ matrix $\mathbf{Z}_i$, which is subsequently processed by a Transformer. As before, the use of the Transformer is motivated by its superior performance in the ``Raw MTS to DL models'' experiments.  This configuration achieved an ROC-AUC of 79.83\%.}  
    
    \item \textcolor{black}{\textit{Hybrid row-column DR + Flattening and MLP:} In this case, we apply two different DR modules. Each feature vector $[\mathbf{X}_i]_{(:,t)}$ is first projected into a latent space via an MLP, generating $\mathbf{e}_{i,t} \in \mathbb{R}^{F^*}$. We then concatenate the $T$ embeddings into the  ${F^* \times T}$ matrix $\mathbf{E}_i$. Secondly, we map each row of $\mathbf{E}_i$ into a lower dimensional space with dimension $T^*<T$ using an MLP. This allows the model to extract intra-time-step patterns before introducing temporal context. The process is repeated for each of the $F^*$ rows, giving rise to a latent matrix representation $\mathbf{Z}_i$ of size $F^*\times T^*$. Matrix $\mathbf{Z}_i$ is then flattened, giving rise to vector $\mathbf{z}_i\in\mathbb{R}^{F^*T^*}$, which is processed using an MLP for its classification. This hybrid design seeks to leverage the complementarity of spatial abstraction and temporal reasoning, offering a balanced modeling perspective. It achieved an ROC-AUC of 77.74\%.} 
    
    \item \textcolor{black}{\textit{Raw data + Flattening and MLP:} In this final configuration, $\mathbf{X}_i \in \mathbb{R}^{F \times T}$ is flattened into a vector $ \mathbf{x}_i \in \mathbb{R}^{FT}$ and processed through a fully connected MLP. While this architecture discards any temporal or spatial structure, it serves to evaluate the contribution of pure feedforward modeling when treated as a tabular classification task. Surprisingly, this simplified representation achieved a competitive ROC-AUC of 79.89\%.}

\end{enumerate}

\textcolor{black}{To ensure a fair comparison across all configurations, models were optimized using consistent hyperparameter search spaces. The learning rate was sampled from the interval $[10^{-5}, 10^{-3}]$, and dropout rates were varied between $0.0$ and $0.3$. For \textit{raw MTS modeling}, recurrent architectures such as LSTM and GRU were evaluated with hidden dimensions selected from $\{10, 20, 30, 35, 40, 45, 50\}$. Moreover, norm-2 regularization was considered for GRU, LSTM, and Transformer architectures, with penalty values set to $\{0.0, 10^{-2}, 10^{-4}\}$. The Transformer model incorporated additional specific parameters, including the number of attention heads $\{2, 4, 6, 8\}$ and stacked blocks $\{2, 4, 6\}$. For \textit{feature-wise DR}, \textit{feature-only processing}, and \textit{hybrid modeling} strategies, the same learning rate and dropout ranges were retained to ensure comparability. Hidden dimensions were explored in the interval $[4, 128]$, while the number of recurrent layers was varied between 1 and 4. When applicable, the embedding dimensionality was chosen from $[4, 128]$, and for PCA-based preprocessing, the number of retained components was selected from $[4, 50]$.}

\textcolor{black}{Figure~\ref{fig:new_results} presents the performance distribution across the five evaluated configurations. Models incorporating input transformations—such as DR or intermediate embeddings—consistently outperform those trained directly on raw data. However, these improvements are sometimes accompanied by increased variability across cross-validation splits. Particularly noteworthy is the architecture that processes each time step independently (feature-only processing), which achieves performance comparable to our interpretable framework. Collectively, these results demonstrate that our proposed approach remains highly competitive with conventional DL models. This underscores its potential for real-world clinical decision support, where both interpretability and predictive reliability are essential.}

\subsection{Clinical knowledge extraction through visualization tools}\label{sec:experiments_visualization_tools}

Recognizing the importance of utilizing MTS analysis techniques, we aim to explore various (MTS-based) visualization tools that enhance interpretability and provide insights. Given the infeasibility of performing this analysis for all 24 schemes examined in the previous section, we focus on a similarity metric based on the DTW$_D$ scheme followed by an exponential kernel, which achieved the highest ROC-AUC in Section~\ref{sec:MDRprediction}.

\begin{figure}[h!]
\centering
	\begin{subfigure}[]
		\centering
        \includegraphics[width=0.3\textwidth]{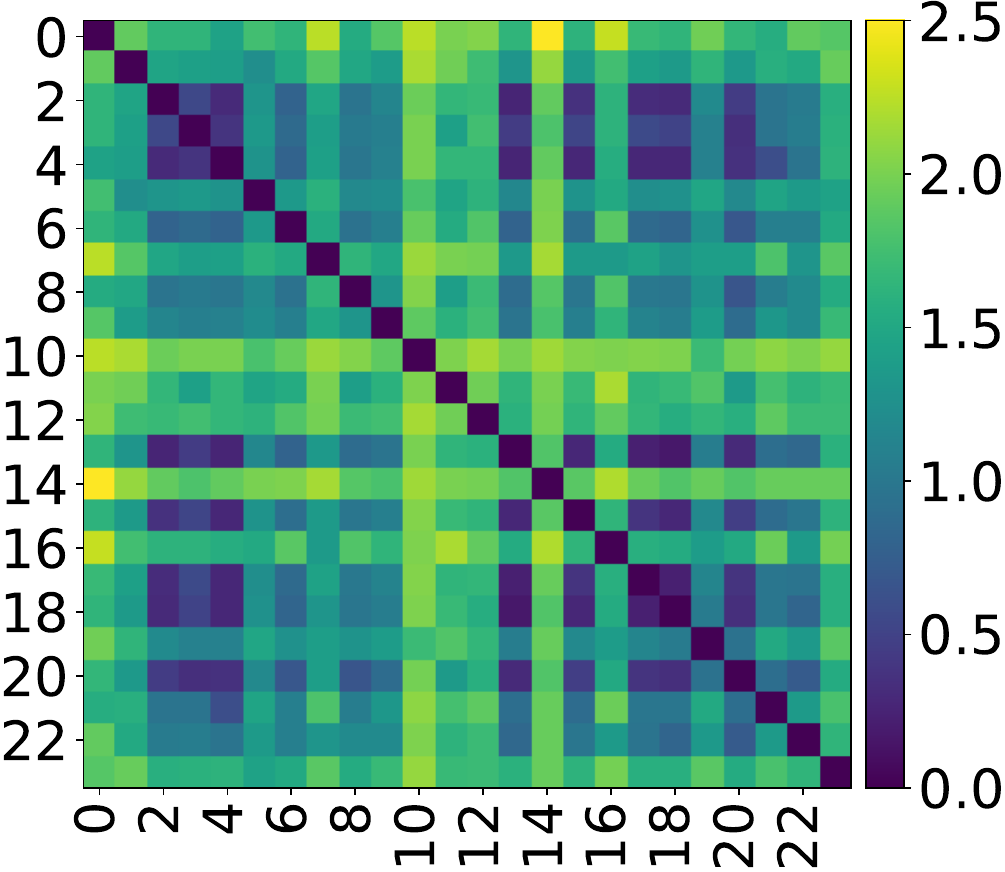}
	\end{subfigure}
	\begin{subfigure}[]
		\centering
        \includegraphics[width=0.3\textwidth]{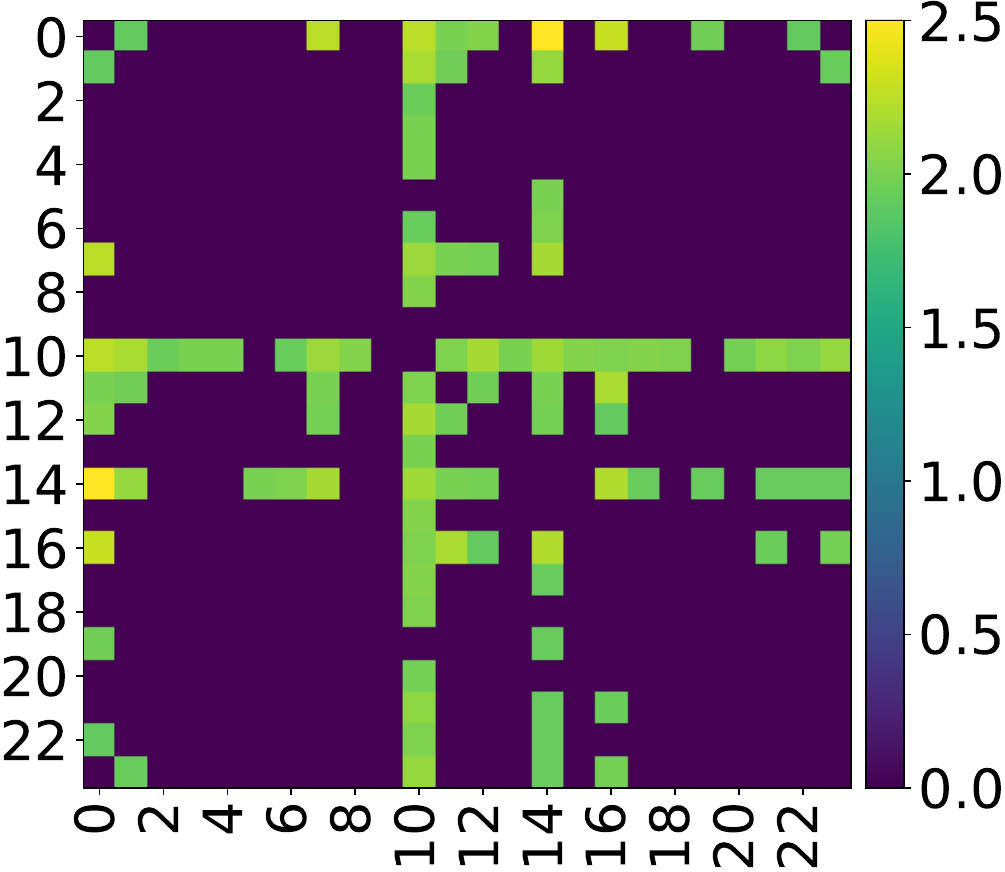}
	\end{subfigure}
    \begin{subfigure}[]
		\centering
        \includegraphics[width=0.3\textwidth]{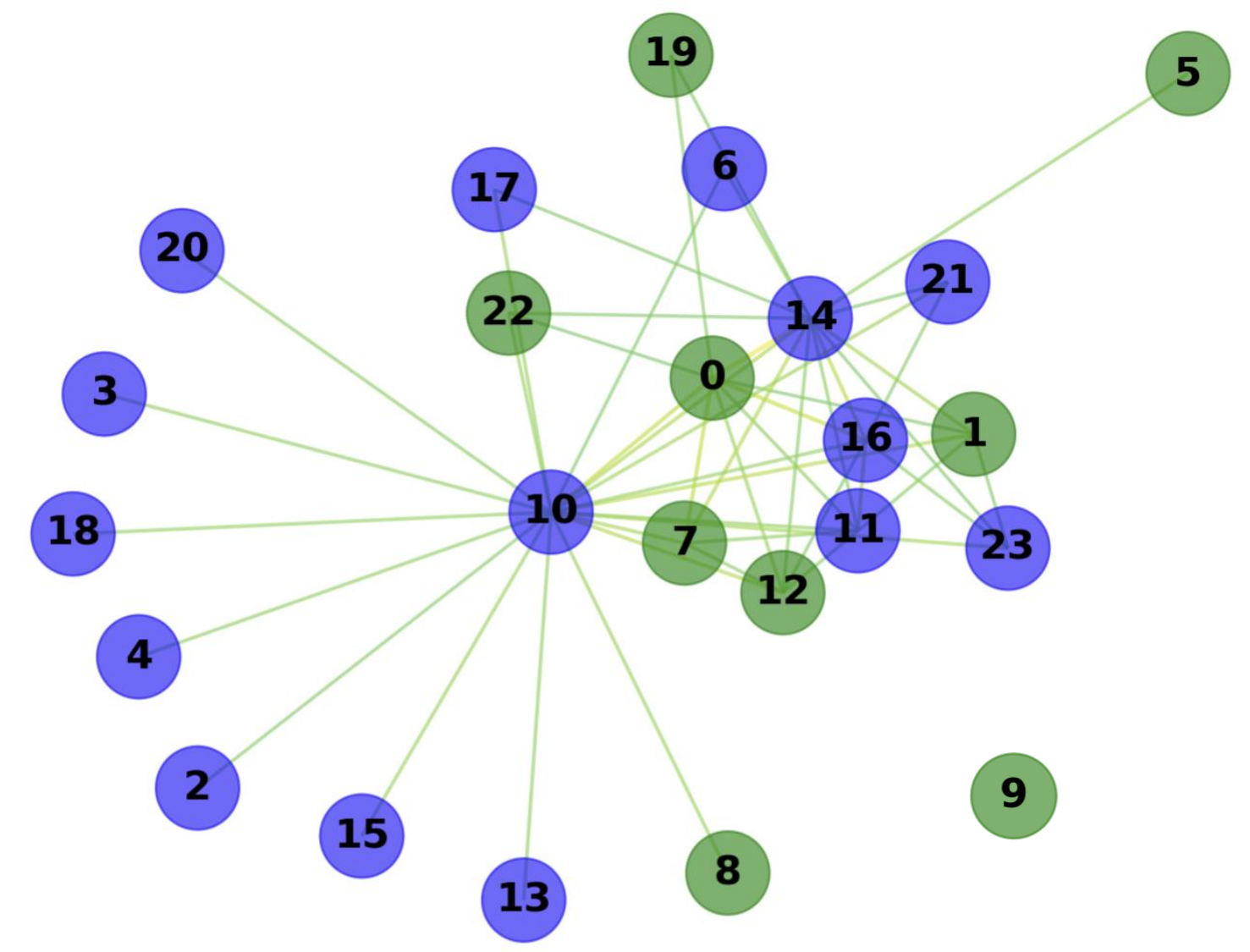}
	\end{subfigure}
	\caption{\textcolor{black}{Representation of the similarity matrix for a $\mathcal{D}_{train}$ dataset acquired through DTW$_D$ combined with an exponential kernel: (a) original similarity matrix; (b) similarity matrix after applying a threshold that removes values below $1.9$; and (c) graphical representation derived from (b), where blue circles represent MDR patients and green circles represent non-MDR patients.}}

	\label{vis_graph}
\end{figure}

The initial step involves constructing a graph in which nodes represent patients, with connections established between patients whose similarity exceeds a specified threshold. Graph-based analysis tools are prevalent in ML and statistics, offering a novel approach to analyzing ICU EHR data, and potentially identifying patterns, relationships, and trends that can aid in the management of MDR patients. Due to the large number of patients in our dataset, instead of representing a comprehensive graph with $842$ patients ($\mathcal{D}_{train}$), we begin by representing a (sub-)graph with the first $24$ patients. Figure~\ref{vis_graph} illustrates the graph construction process, showing: (a) the $24\times 24$ similarity matrix where entry $(i,j)$ represents the similarity between the MTS associated with patients $i$ and $j$ (left panel); (b) a sparse similarity matrix where relations below a threshold are set to zero (central panel); and (c) an unweighted graph built using the sparse matrix in (b) (right panel). In the graph, patients (nodes) who developed MDR are represented by green circles, while non-MDR patients are indicated by blue circles.

It is important to note that static variables were not used in constructing the predictive models or for the clustering process. Analyzing the graph in Figure~\ref{vis_graph}(c), two distinct subgraphs are apparent: i) a star graph (with patient 10 as the central node) primarily consisting of MDR patients (top part of the graph) and ii) a more densely connected subgraph comprising both MDR and non-MDR patients (bottom part of the graph). Notably, the nodes in the top (star) subgraph correspond to patients who developed MDR within the initial $48$ hours of ICU admission. These patients, with an average age of approximately 70 years, were in deteriorating health. In contrast, the nodes in the lower subgraph represent patients who: i) spent several days in the ICU (MDR patients developed MDR after an average of $5.25$ days, while non-MDR patients stayed for $6.33$ days) and ii) were exposed to high antibiotic pressure. Additionally, peripheral nodes $5$ and $19$ correspond to younger patients with overall good health, who did not require MV and may have extended ICU stays.

While the qualitative analysis of smaller graphs provides valuable insights, extending this analysis to larger graphs becomes increasingly complex. To address this, we implement systematic similarity/graph-based visualization and knowledge extraction tools that automate the process. Specifically, we employ a spectral clustering graph-based method to automatically group nodes and the t-SNE methodology to visualize these clusters (see Sections~\ref{subsec-tsnemethods} and~\ref{sc} for technical details). This visual representation has the potential to reveal insights that: (i) contribute to characterizing the appearance of MDR and (ii) might not have been readily discernible within the original multidimensional space. Due to page-limit constraints, we present results for a subset of the 24 combinations analyzed in Section~\ref{sec:MDRprediction}. The full set of representations is available on the GitHub repository:~{\small\url{https://github.com/oscarescuderoarnanz/DM4MTS/tree/main/Step2_Clustering_and_graphs/Visual_2D}}. 
This subset was selected based on dimensions such as: i) differences among the approaches considered; ii) usefulness in detecting meaningful patient patterns with clinical utility; and iii) the best ROC-AUC (median) results in classification experiments.

\begin{figure}[h!]
\centering
	\begin{subfigure}[]
		\centering
		\includegraphics[width=0.23\textwidth]{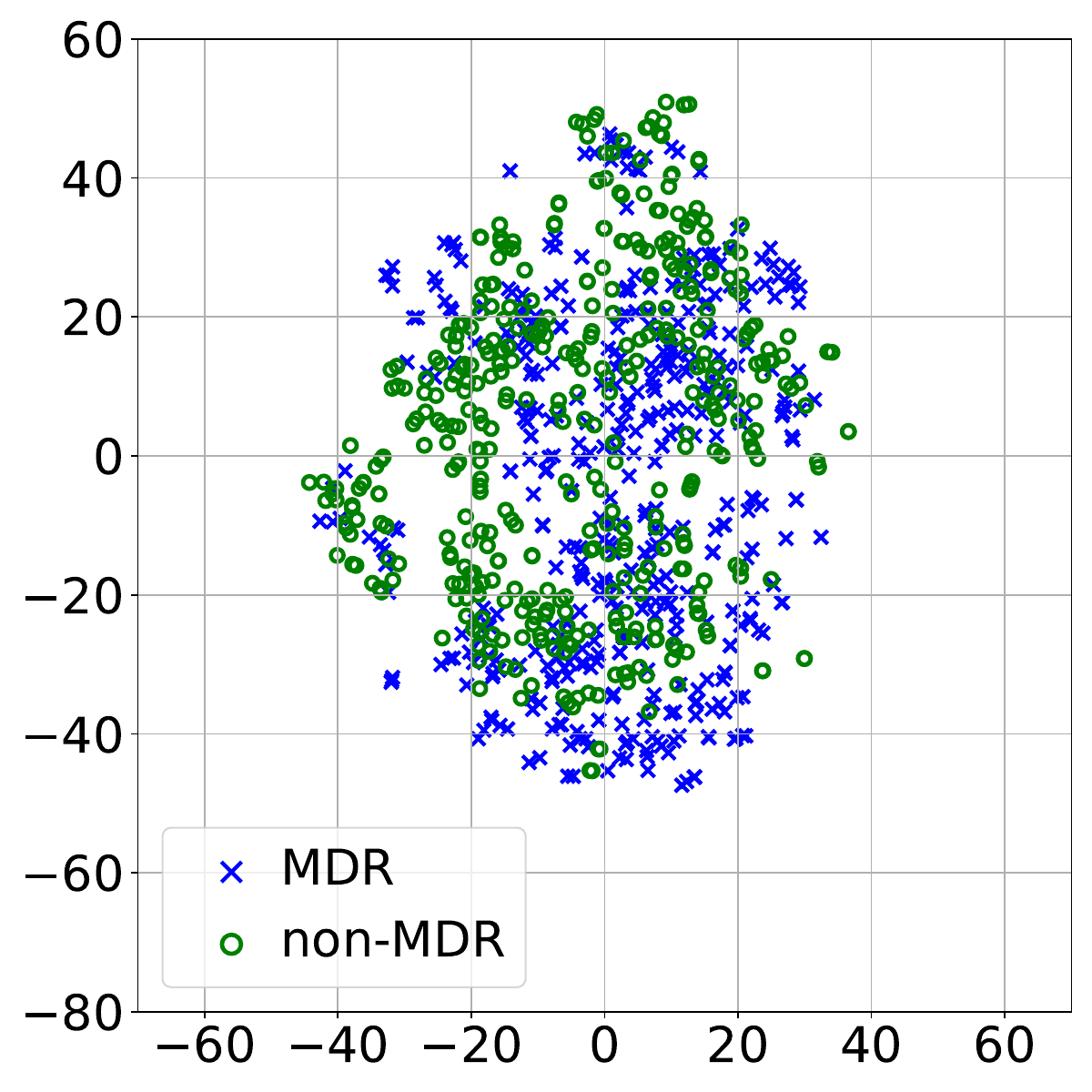}
	\end{subfigure}
	\begin{subfigure}[]
		\centering
		\includegraphics[width=0.23\textwidth]{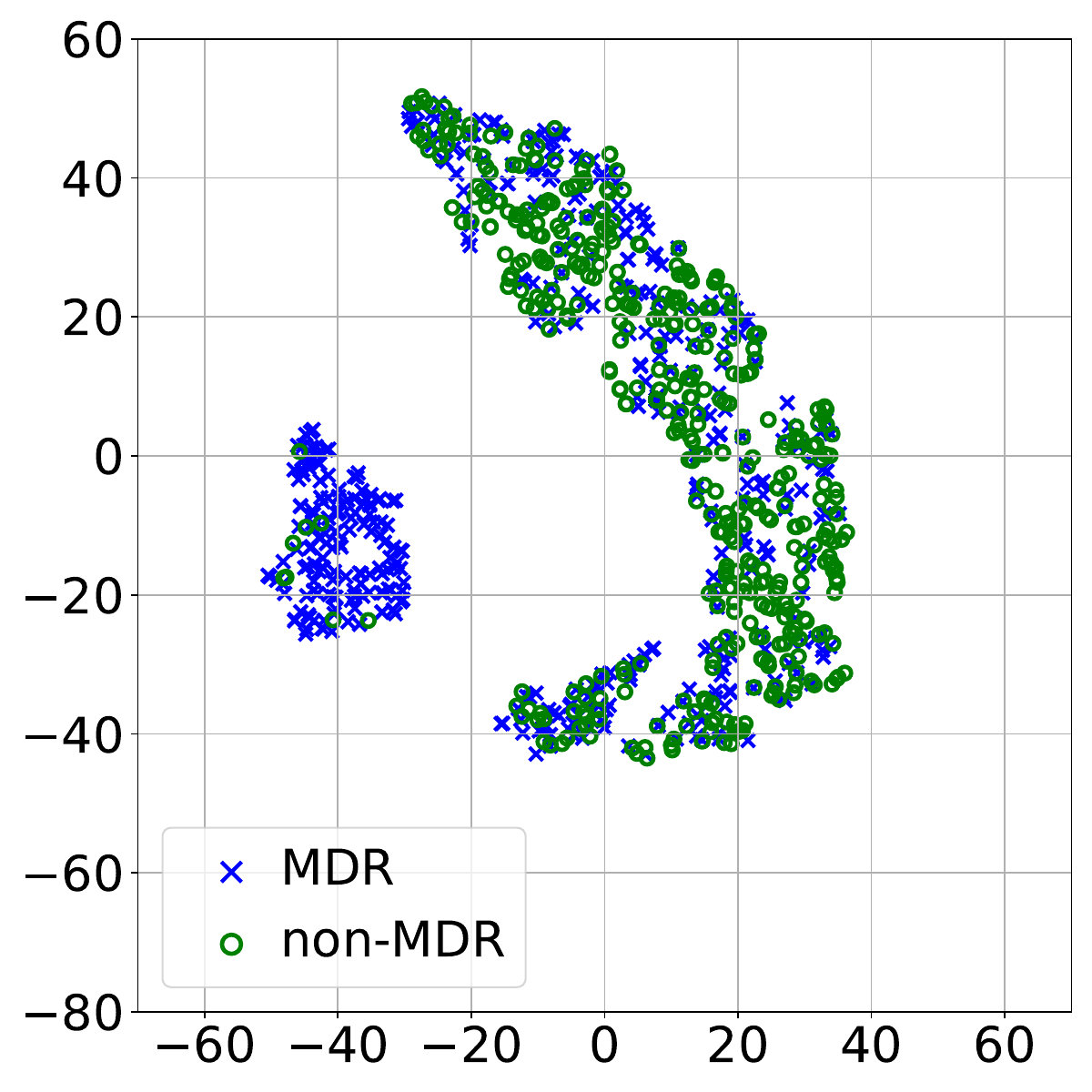}
	\end{subfigure}
	\begin{subfigure}[]
		\centering
		\includegraphics[width=0.23\textwidth]{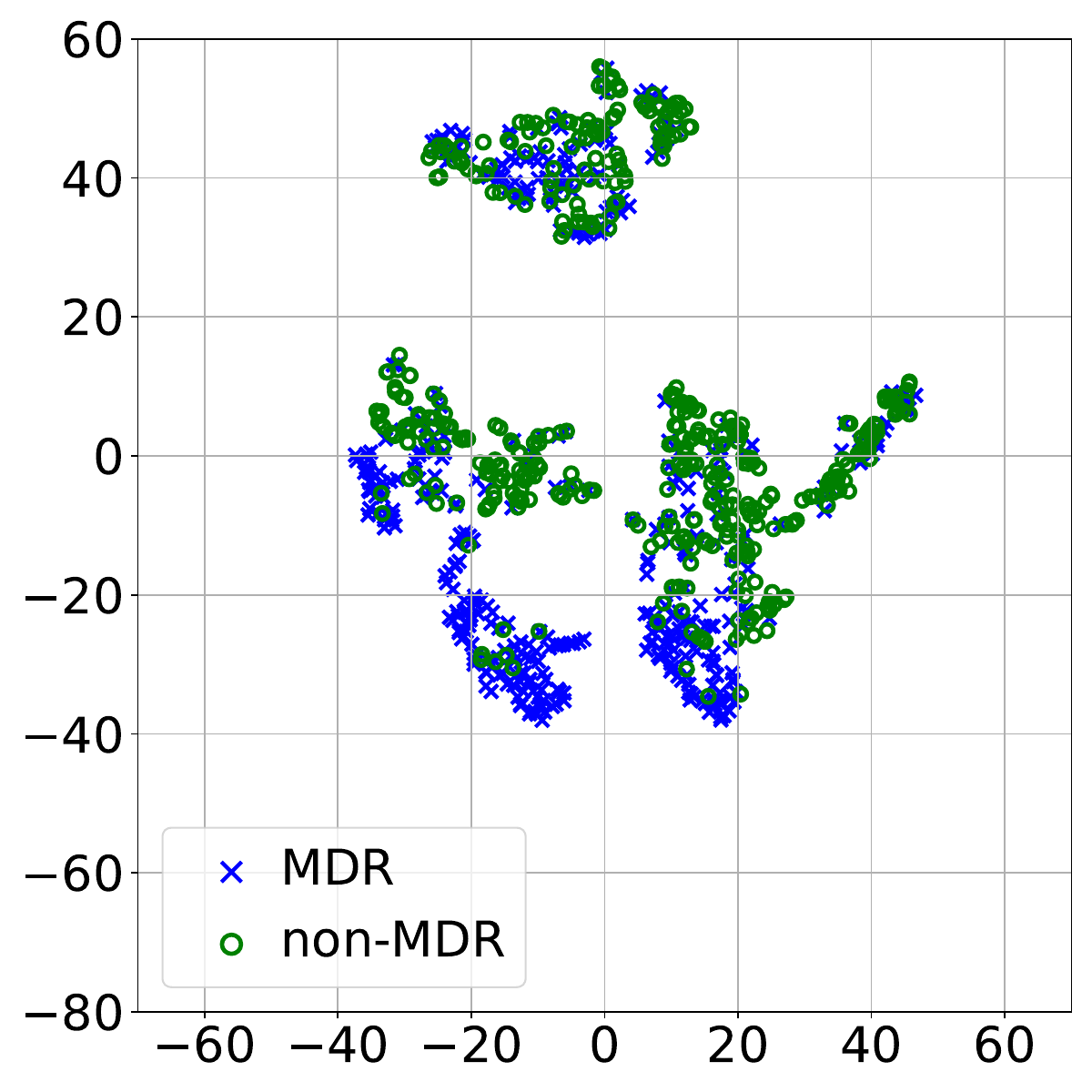}
	\end{subfigure}
		\begin{subfigure}[]
		\centering
		\includegraphics[width=0.23\textwidth]{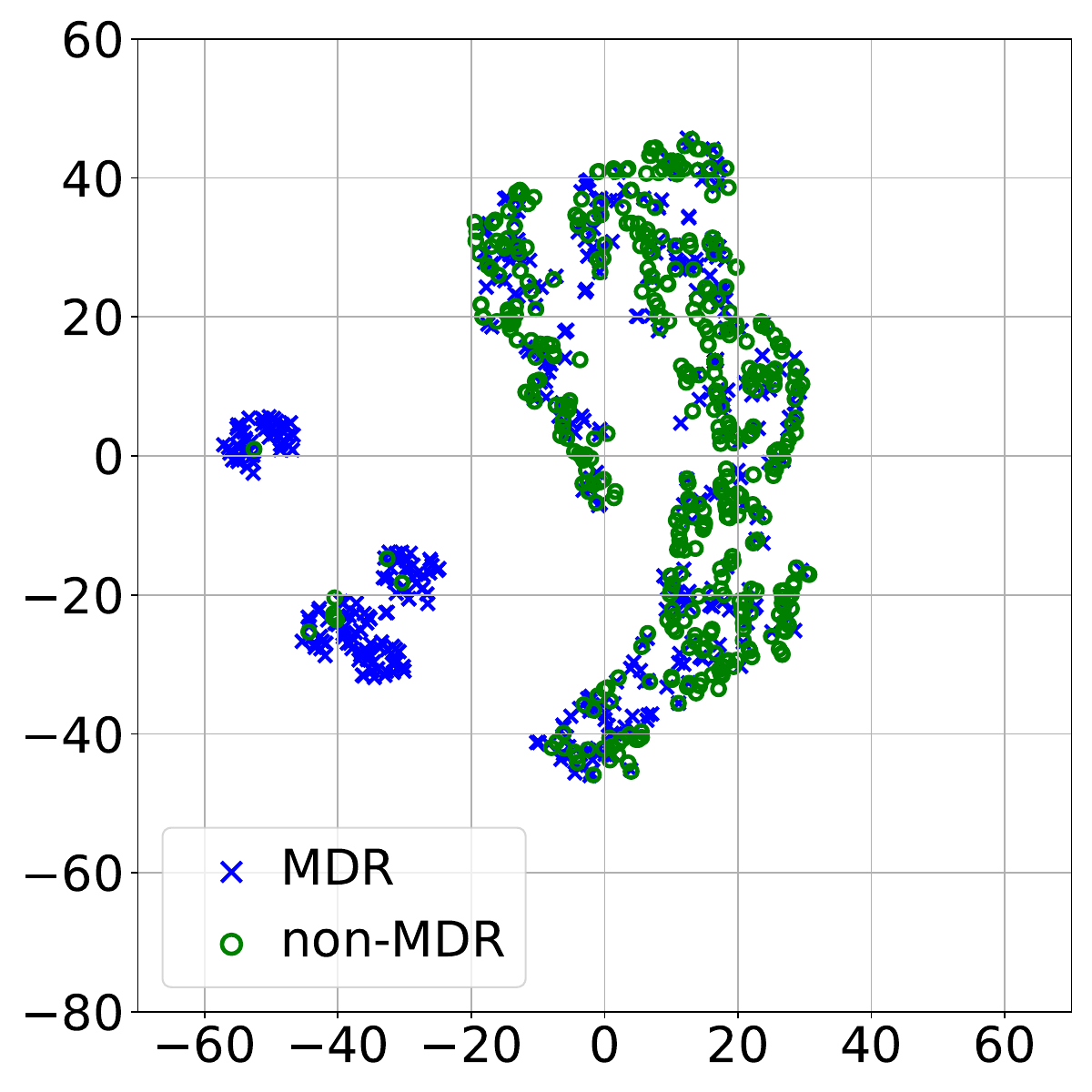}
	\end{subfigure}

	\caption{Projection of patient data onto a 2D space using t-SNE after modeling the MTS with: (a) FE and PCA; (b) DTW$_D$ and exponential kernel; (c) DTW$_I$ and AE; (d) TCK without DR.}
	\label{someVisualizations_TSNE}
\end{figure}

In Figure~\ref{someVisualizations_TSNE}, 2D embeddings generated by t-SNE after processing MTS data using four different techniques are displayed: FE-PCA (panel a), DTW$_D$ with exponential kernel (panel b), DTW$_I$-AE (panel c), and TCK without DR (panel d). Patients with MDR and those without are distinguished by unique markers and colors (green circles for non-MDR patients and blue crosses for MDR patients). The combination of a straightforward FE scheme and linear PCA results in a single cluster with scattered points, failing to reveal meaningful patterns. This trend is consistent across all combinations with FE, reinforcing the idea that methods accounting explicitly for the MTS nature of the data yield superior results. In contrast, the remaining MTS-based combinations provide valuable clinical insights for characterizing MDR behavior. Consequently, we focus on combinations that have demonstrated the best ROC-AUC (median) performance in classification: i) DTW$_D$ with kernel transformation and ii) TCK without DR.

\begin{figure}[h!]
\centering
	\begin{subfigure}[]
		\centering
		\includegraphics[width=0.475\textwidth]{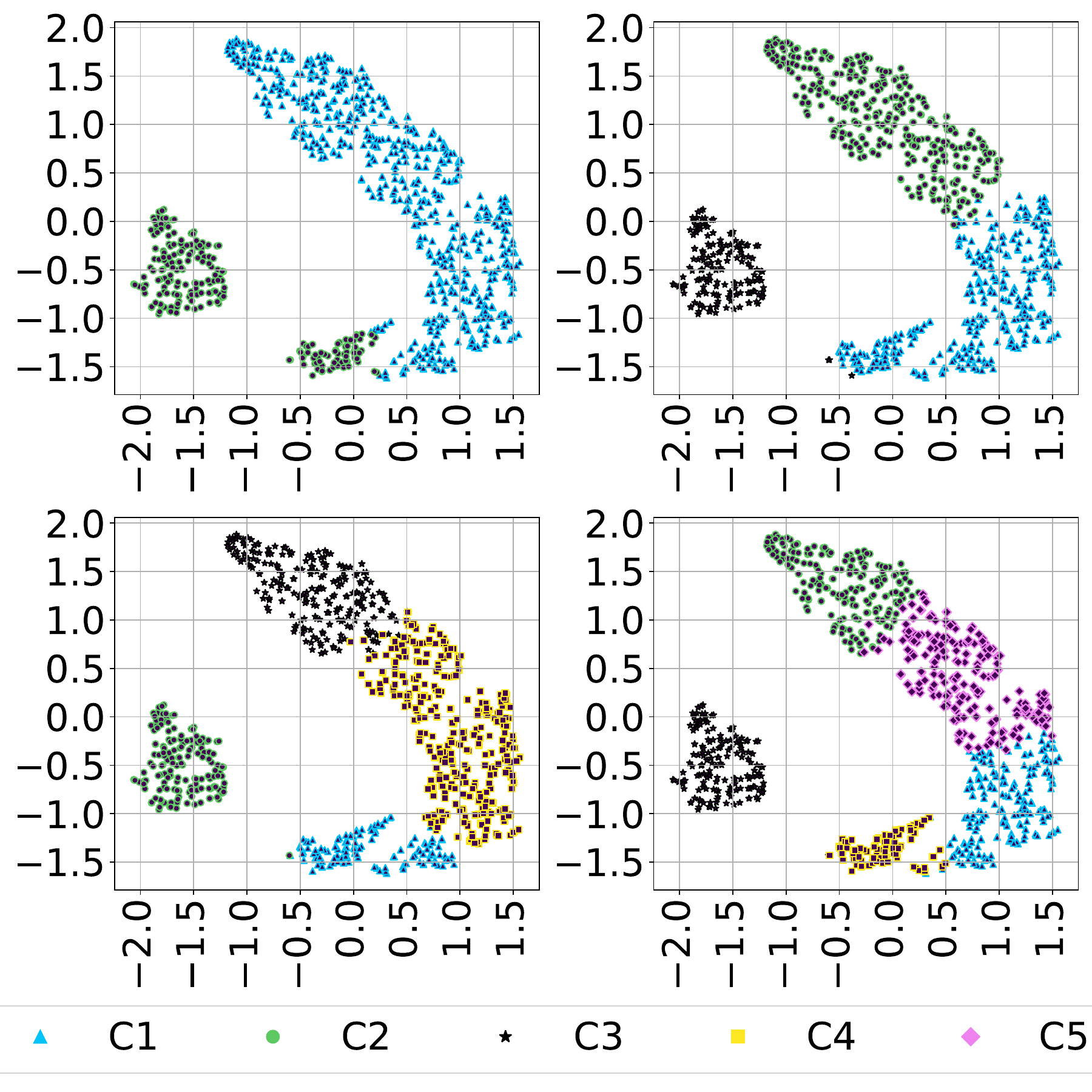}
	\end{subfigure}
	\begin{subfigure}[]
		\centering
		\includegraphics[width=0.475\textwidth]{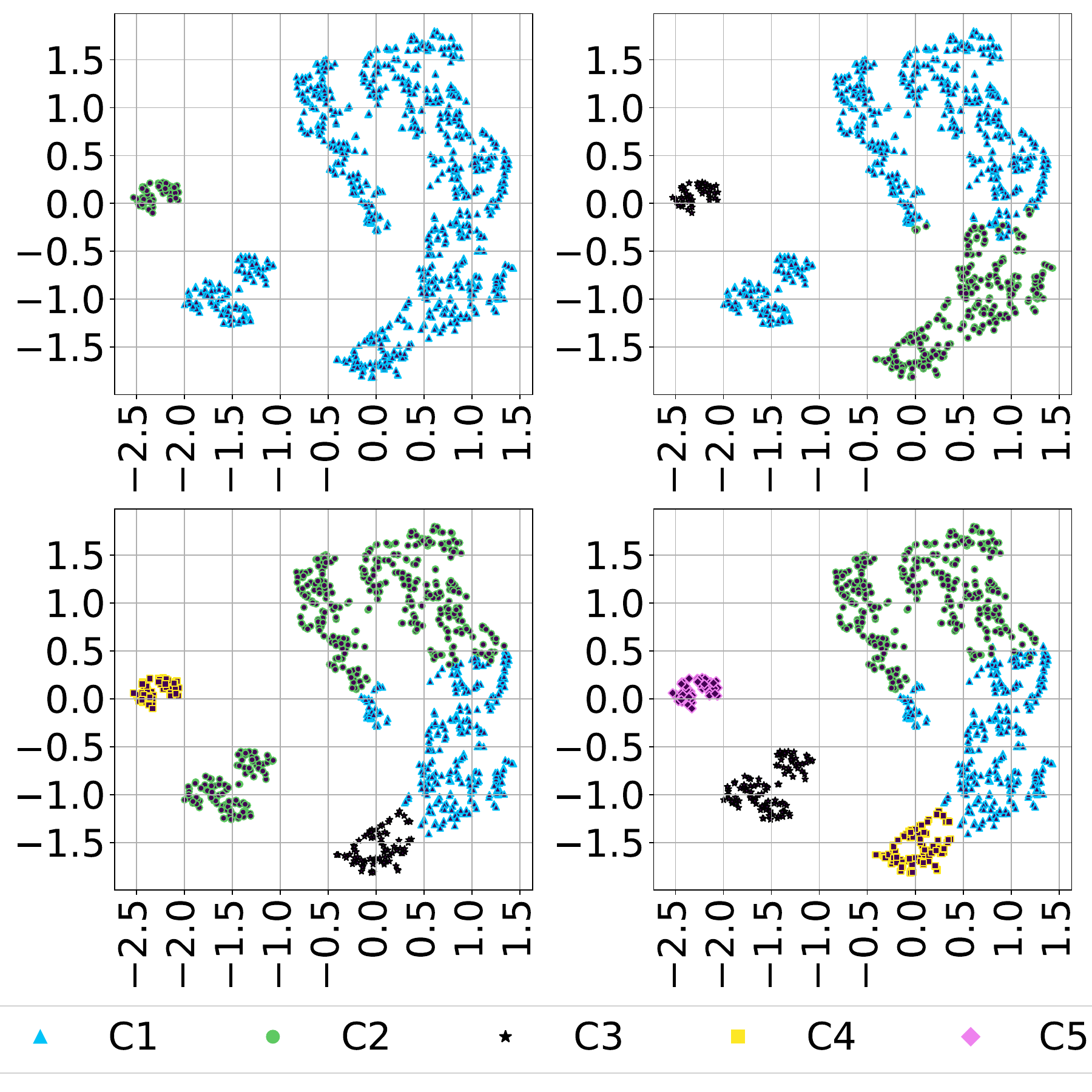}
	\end{subfigure}

        \caption{\textcolor{black}{Spectral clustering (with a different number of clusters, from 2 to 5) applied to MTS modeling using: (a) DTW with exponential kernel and (b) TCK without DR, combined with t-SNE. Legends are provided below each figure (a) and (b).}}

	\label{spectralClustering}
\end{figure}

To conduct a more rigorous assessment, we apply a spectral clustering algorithm to the representations in Figure~\ref{someVisualizations_TSNE} (b) and (d). These representations were identified by clinical experts as the most relevant and exhibited the best classification performance in terms of median ROC-AUC. As explained in Section~\ref{visualization}, the graph-based spectral clustering method seeks to partition the dataset into non-overlapping subsets, ensuring that data assigned to different subsets are dissimilar. Figure~\ref{spectralClustering} displays the clusters obtained for different values of $C$, where $C$ denotes the number of clusters. The upper panels correspond to $C=2$ and $C=3$, while the lower panels correspond to $C=4$ and $C=5$. The $i$-th cluster is indicated in the legend as $C_i$.

\begin{figure}[h!]
\centering
	\begin{subfigure}[]
		\centering
        \includegraphics[width=0.235\textwidth]{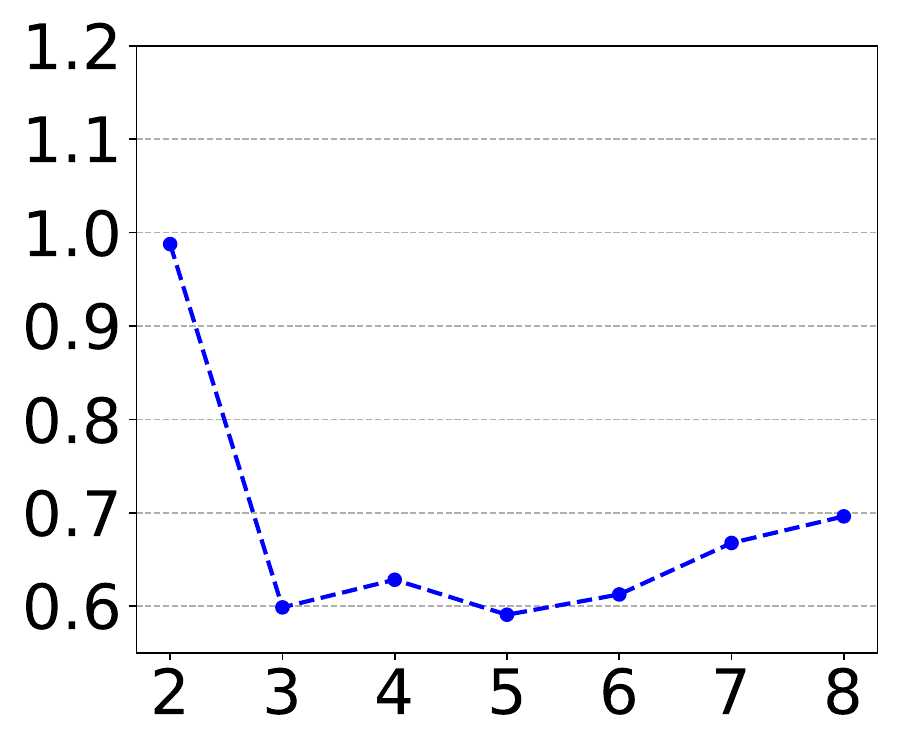}
	\end{subfigure}
	\begin{subfigure}[]
		\centering
        \includegraphics[width=0.235\textwidth]{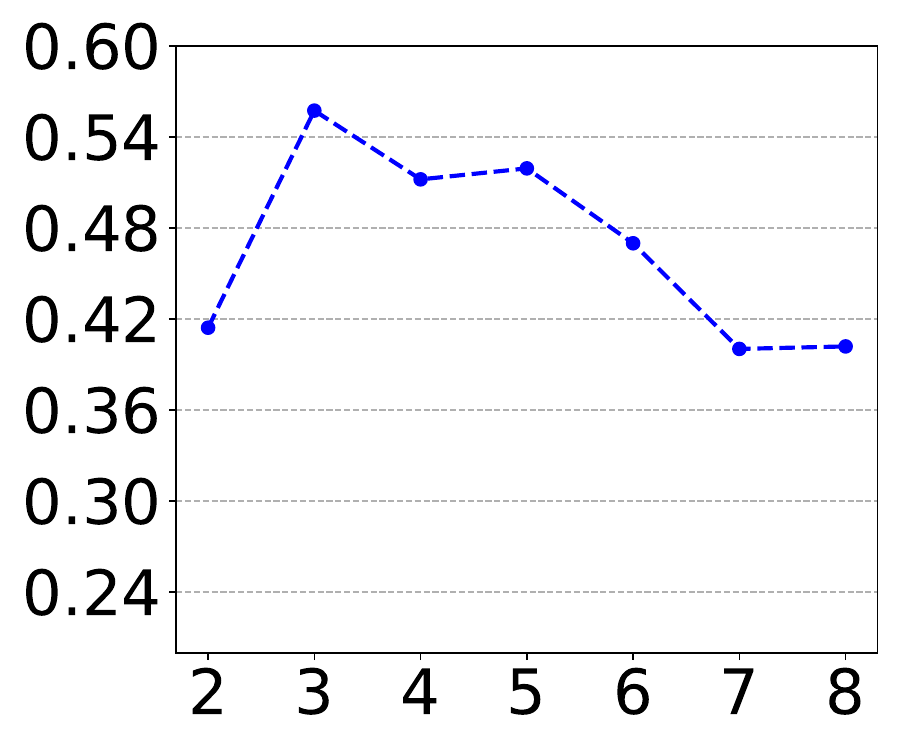}
	\end{subfigure}
    \begin{subfigure}[]
		\centering
        \includegraphics[width=0.235\textwidth]{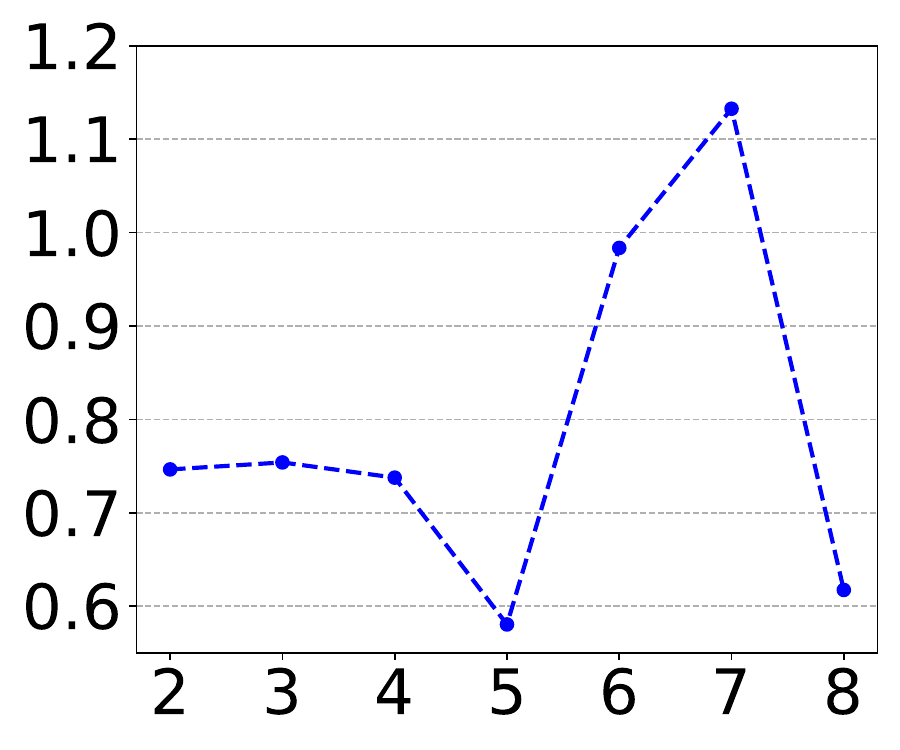}
	\end{subfigure}
    \begin{subfigure}[]
		\centering
        \includegraphics[width=0.235\textwidth]{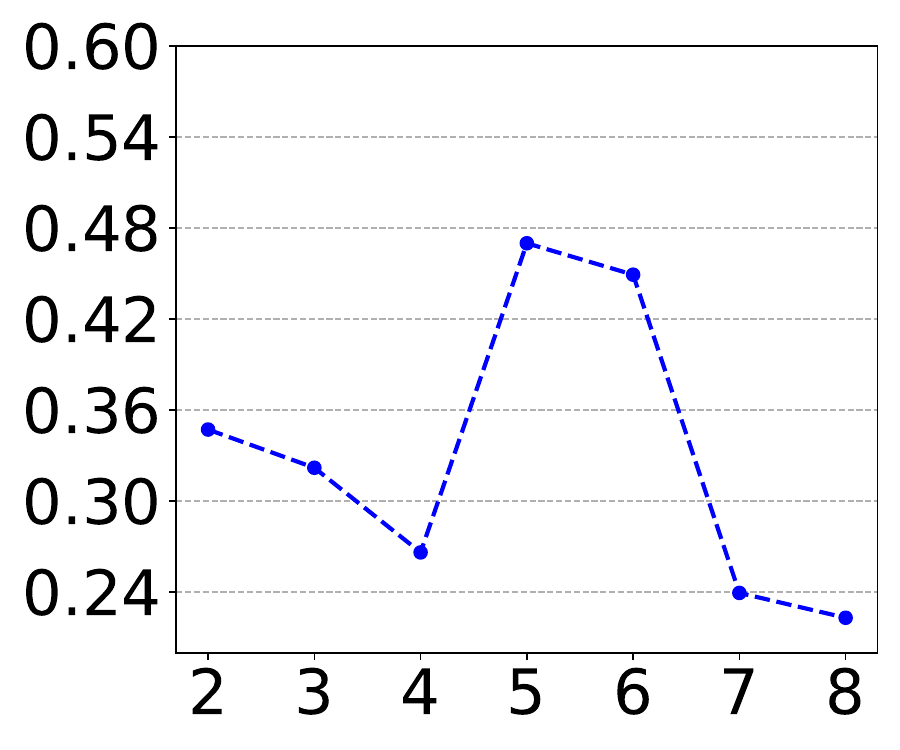}
	\end{subfigure}
    \vspace{-0.3cm}
	\caption{CVIs for varying values of the clustering parameter $C$. (a) and (c) depict the Davis-Bouldin index (ideally lower), while (b) and (d) illustrate the Silhouette index (ideally higher) for DTW and TCK methods, respectively. Both indices consistently suggest that selecting $C=5$ is the optimal choice for the number of clusters.}
	\label{CVIs}
\end{figure}

To determine the optimal number of clusters, we use two CVIs: (a) the Davis-Bouldin index and (b) the Silhouette index (see Section~\ref{visualization} for details). Figure~\ref{CVIs} presents the values of these two CVIs for various values of $C$, both for DTW (Figures~\ref{CVIs}~(a) and (b)) and for TCK (Figures~\ref{CVIs}~(c) and (d)). For the Davis-Bouldin index, the goal is to identify the value of $C$ that yields the minimum CVI. For the Silhouette index, we aim to select the value that results in the maximum CVI. The results from both indicators converge for TCK, establishing $C=5$ as the optimal number of clusters. For DTW, the Davis-Bouldin criterion suggests $C=5$, while the Silhouette index indicates a primary maximum at $C=3$ and a secondary maximum at $C=5$. To ensure consistency in the study and facilitate fair comparisons, we set $C=5$ for both DTW and TCK. Subsequently, a detailed analysis of the resulting clusters was performed separately for both methods.

We begin by analyzing the cluster sizes and the proportion of patients with MDR versus those without in each cluster. Table~\ref{Tabla:PatNum} presents these results. For DTW, a relatively homogeneous distribution is observed among four out of five clusters, except for one cluster with less than 10\% of the patients. Additionally, a cluster predominantly composed of MDR patients is particularly noticeable. In contrast, the TCK method reveals a non-uniform distribution, with two clusters comprising nearly 70\% of the patient population, while the remaining three contain less than 30\%. Notably, two of the five clusters identified by TCK ($C_3$ and $C_5$) consist predominantly of MDR patients, with these two clusters being the smallest in terms of patient numbers. Interestingly, patients in clusters $C_3$ and $C_5$ with TCK are grouped together into cluster $C_3$ when using DTW. This observation led to identifying a patient who, within the TCK framework, was absent from these clusters but belonged to cluster $C_4$. This insight highlights the differences and similarities between the two analysis techniques, emphasizing the importance of understanding how patients are categorized within each method.

\begin{table}[h!]
    \caption{By row is indicated i) the number of patients per group; ii) relative size, considering the total number of patients for a $\mathcal{D}_{train}$ is 842; and iii) the percentage of patients with MDR in each group.}
    \label{Tabla:PatNum}
    \centering
    \begin{tabular*}{\textwidth}{@{\extracolsep\fill}c|cccccc}
    \toprule
    Method &  & $C_1$ &  $C_2$ &  $C_3$ & $C_4$ & $C_5$ \\
    \midrule
    \multirow{3}{*}{DTW} 
    & \# patients &  191  &  198 &  155  &  81  &  217  \\
    & (\%patients) &  (22.68\%)  &  (23.52\%)  &   (18.41\%)  &   (9.62\%)  &  (25.77\%)  \\
    & \% of MDR & 31.41\% & 40.40\% & 94.84\% & 31.41\% &  39.63\%\\
    \midrule
    \multirow{3}{*}{TCK} 
    & \# patients &  281  & 326  &  107  &  81  &  47  \\
    & (\%patients) &  (33.37\%)  &  (38.72\%)  &  (12.71\%)  &   (9.62\%)  &  (5.58\%)  \\
    & \% of MDR & 32.03\% & 41.41\% & 93.46\% & 61.63\% & 61.63\% \\
    \bottomrule
    \end{tabular*}
\end{table}

\begin{figure}[t!]
\centering
    \begin{subfigure}[]
		\centering
        \includegraphics[width=0.18\textwidth]{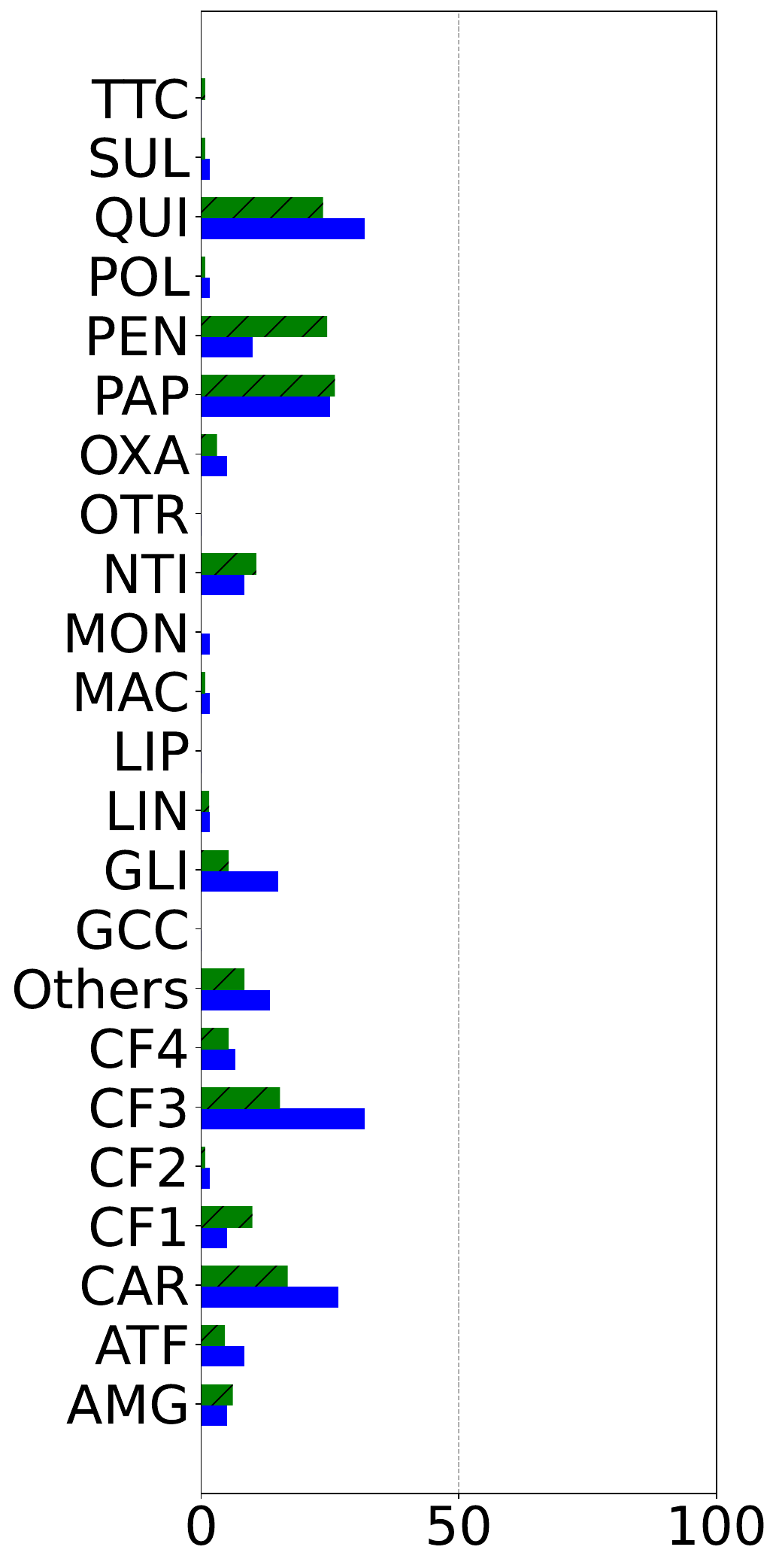}		
    \end{subfigure}
	\begin{subfigure}[]
		\centering
      \includegraphics[width=0.18\textwidth]{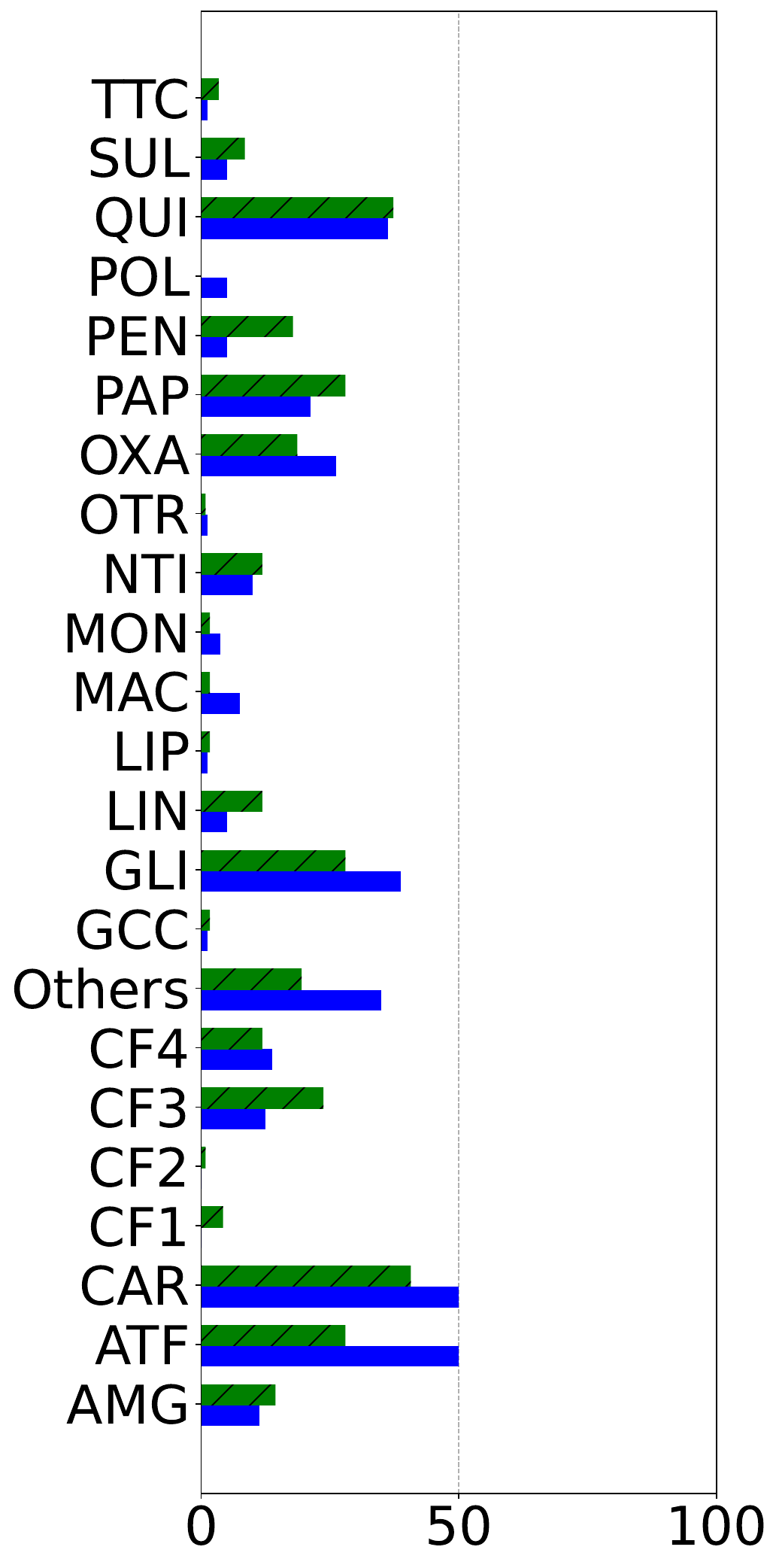}
	\end{subfigure}
	\begin{subfigure}[]
		\centering
        \includegraphics[width=0.18\textwidth]{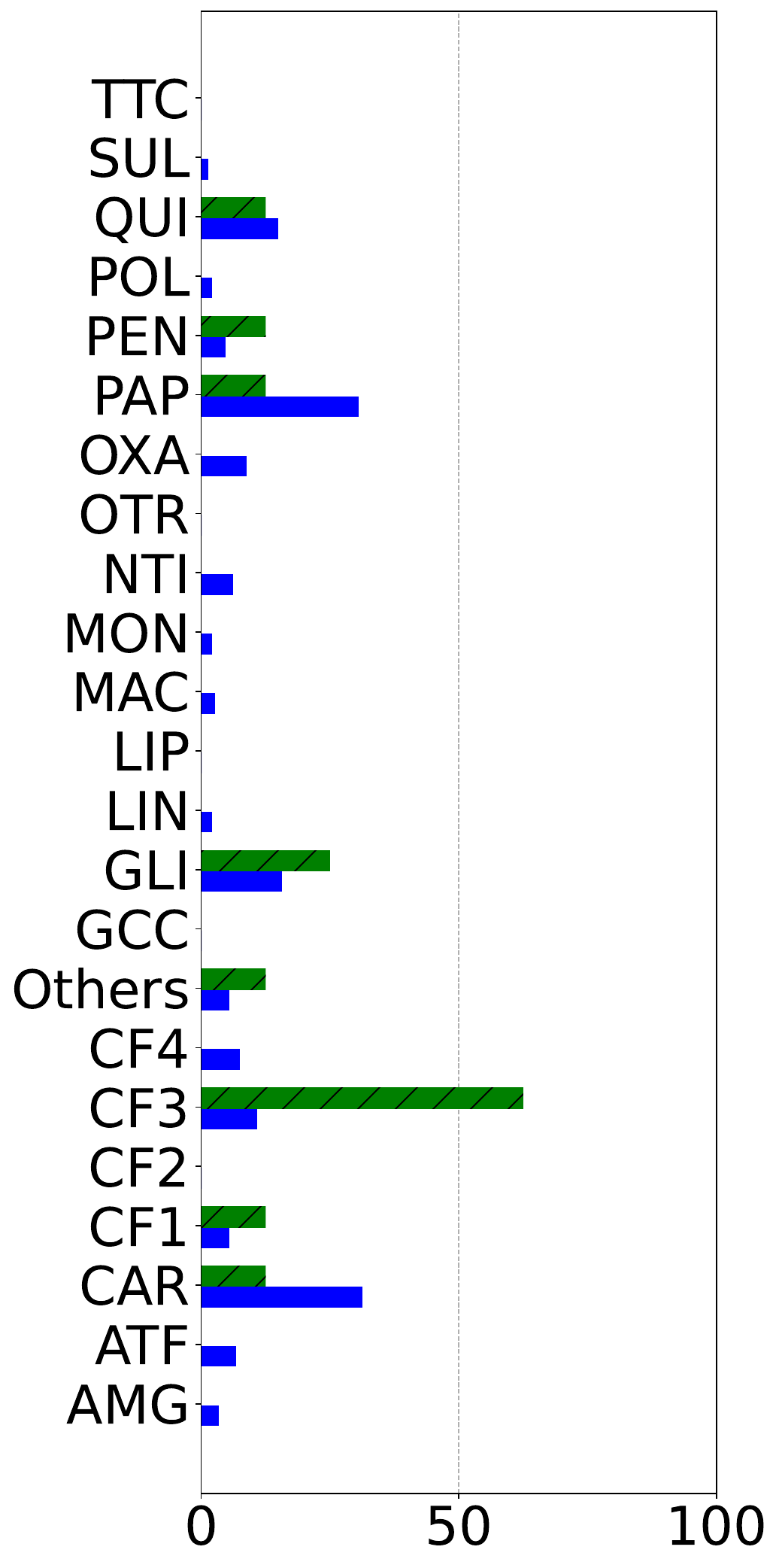}
	\end{subfigure}
	\begin{subfigure}[]
		\centering
         \includegraphics[width=0.18\textwidth]{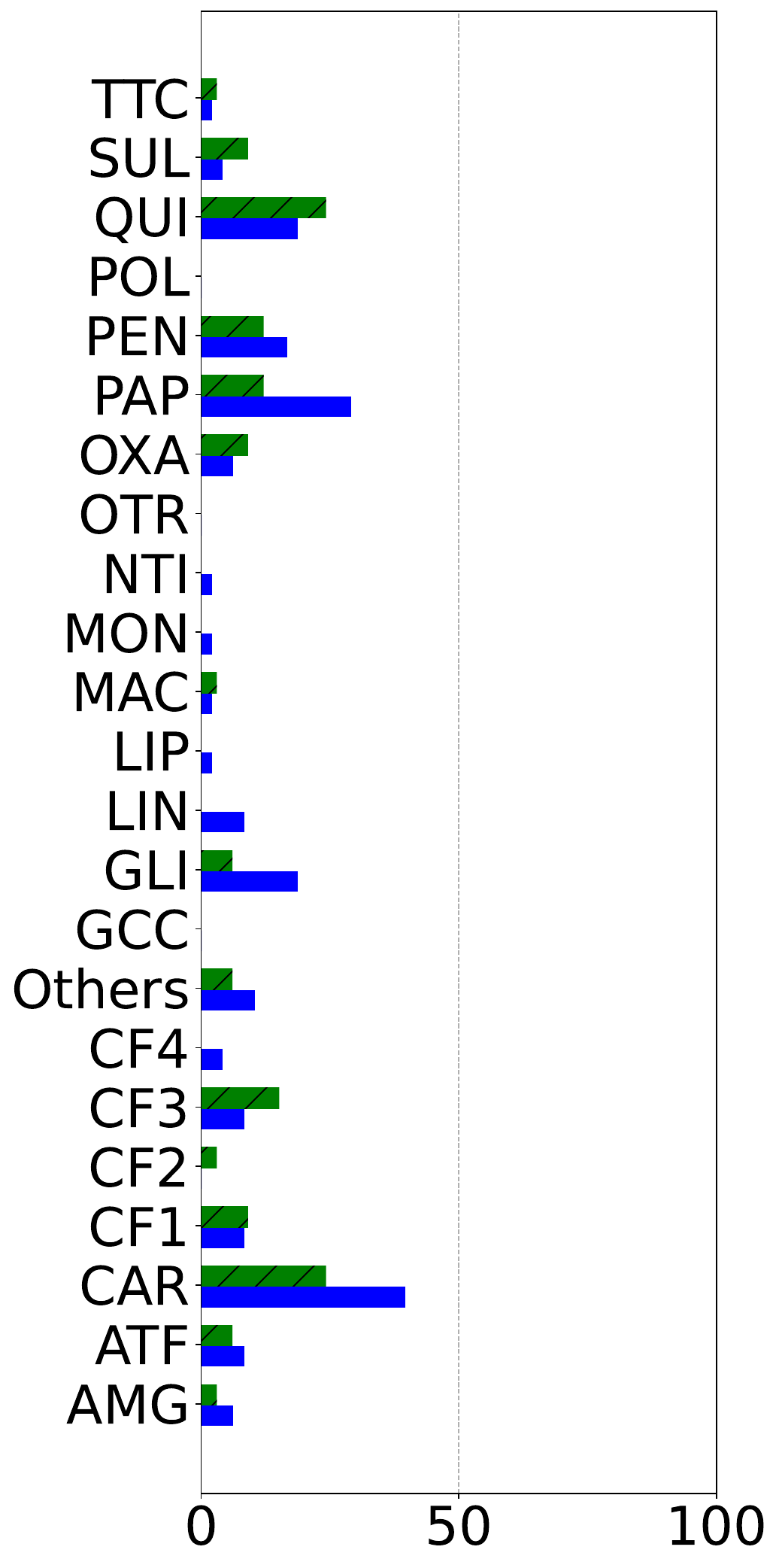}
	\end{subfigure}
	\begin{subfigure}[]
		\centering
         \includegraphics[width=0.18\textwidth]{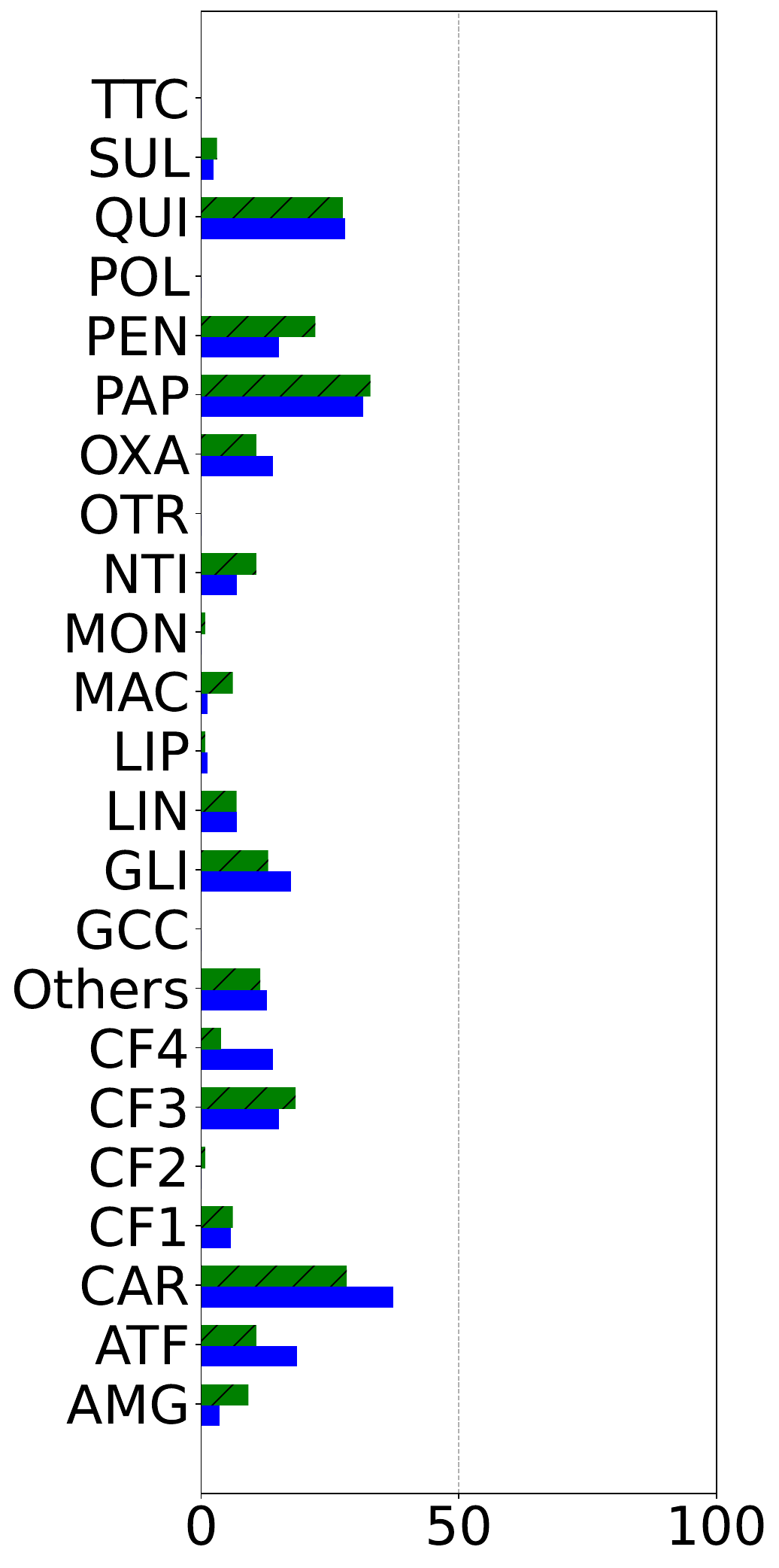}
	\end{subfigure}

 \begin{subfigure}[]
		\centering
        \includegraphics[width=0.18\textwidth]{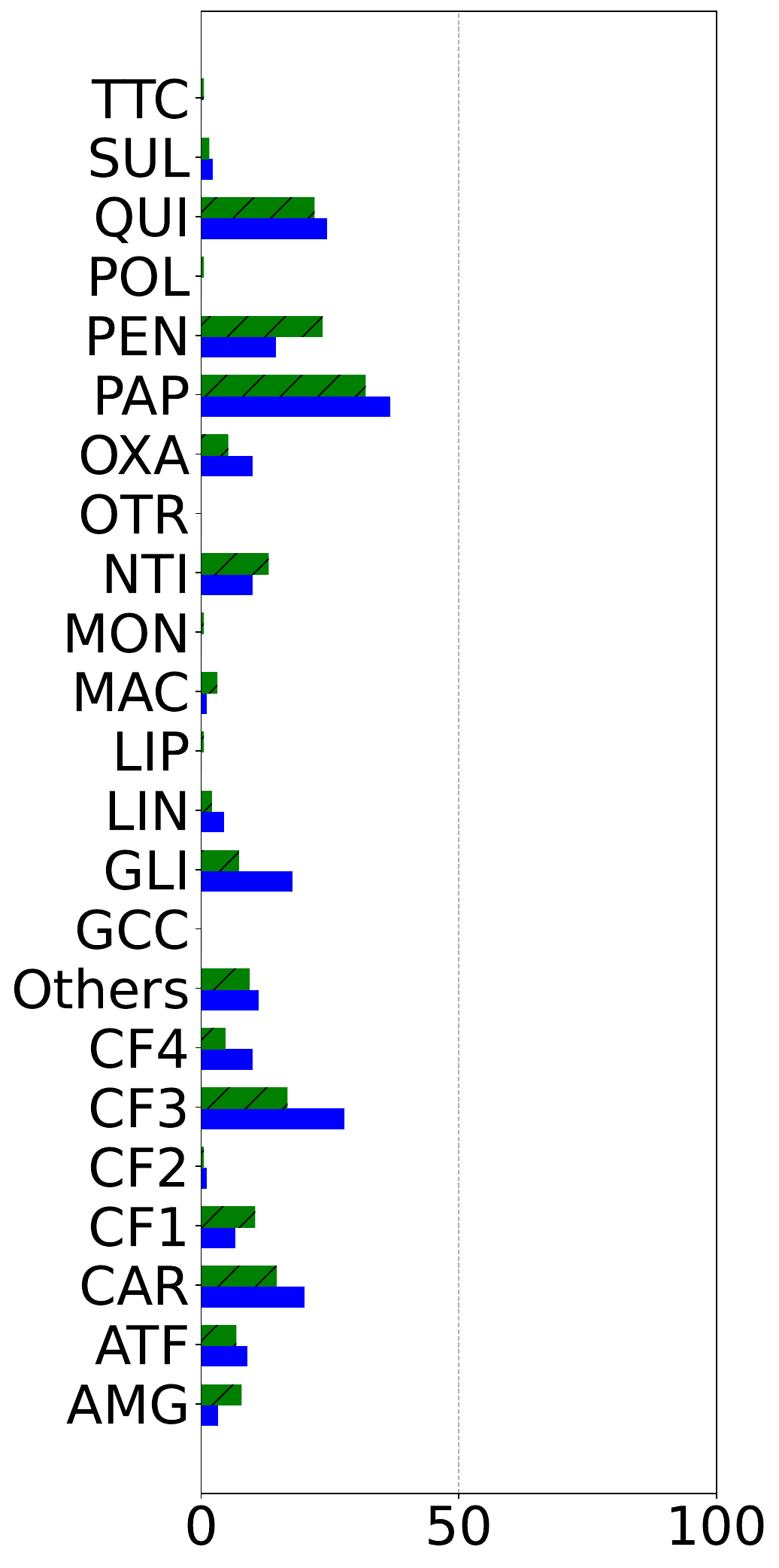}		
    \end{subfigure}
	\begin{subfigure}[]
		\centering
      \includegraphics[width=0.18\textwidth]{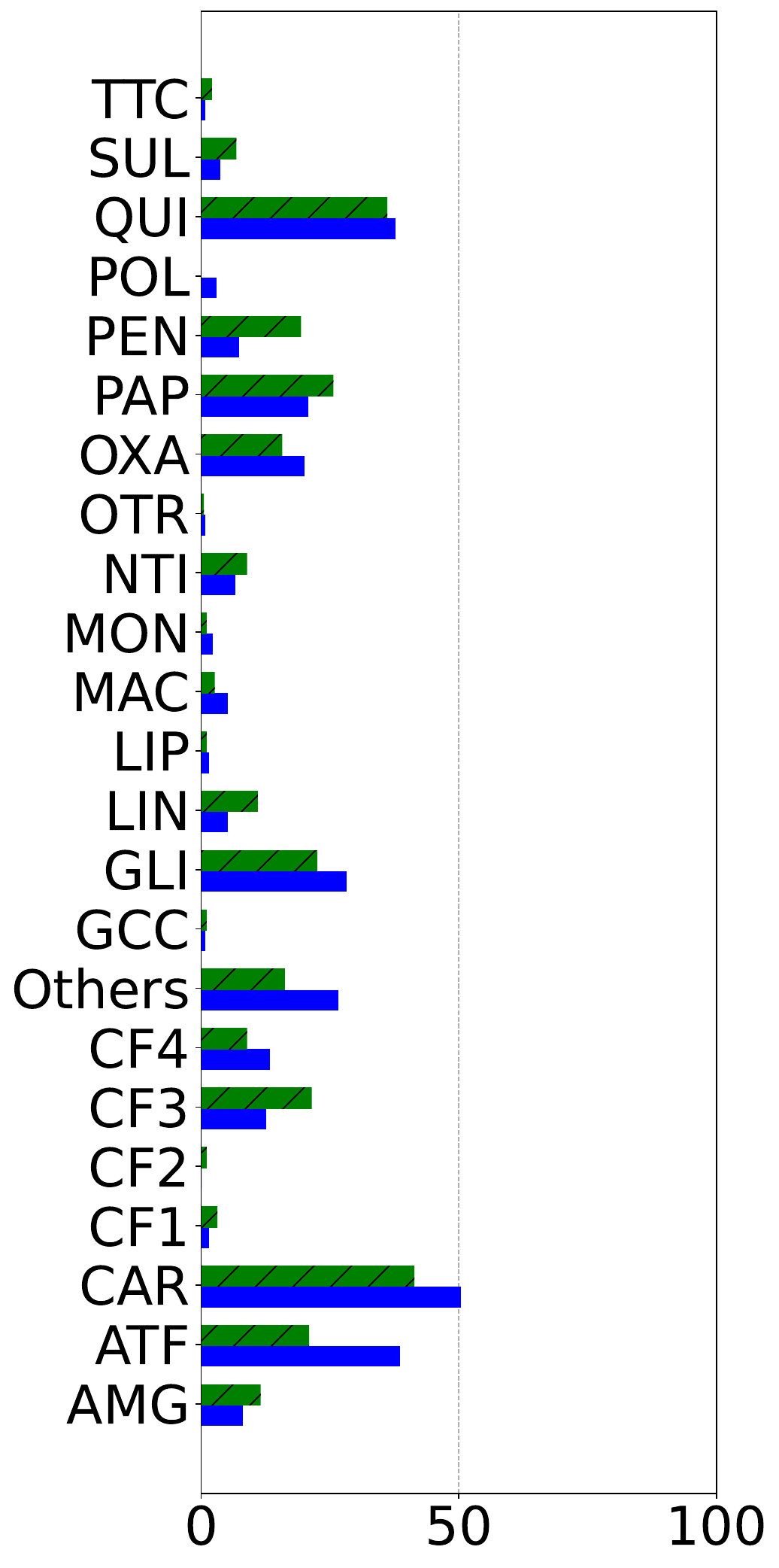}
	\end{subfigure}
	\begin{subfigure}[]
		\centering
        \includegraphics[width=0.18\textwidth]{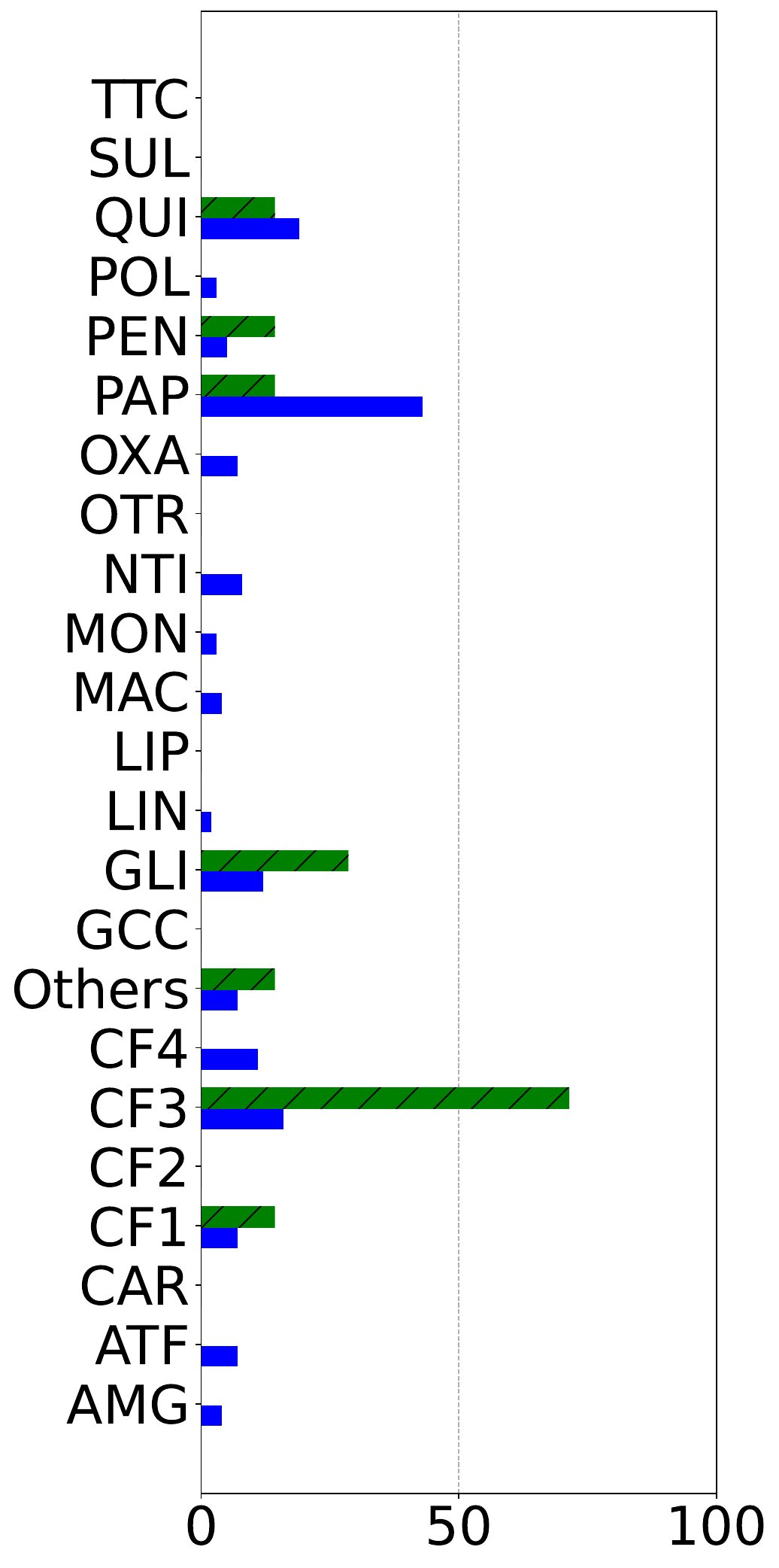}
	\end{subfigure}
	\begin{subfigure}[]
		\centering
         \includegraphics[width=0.18\textwidth]{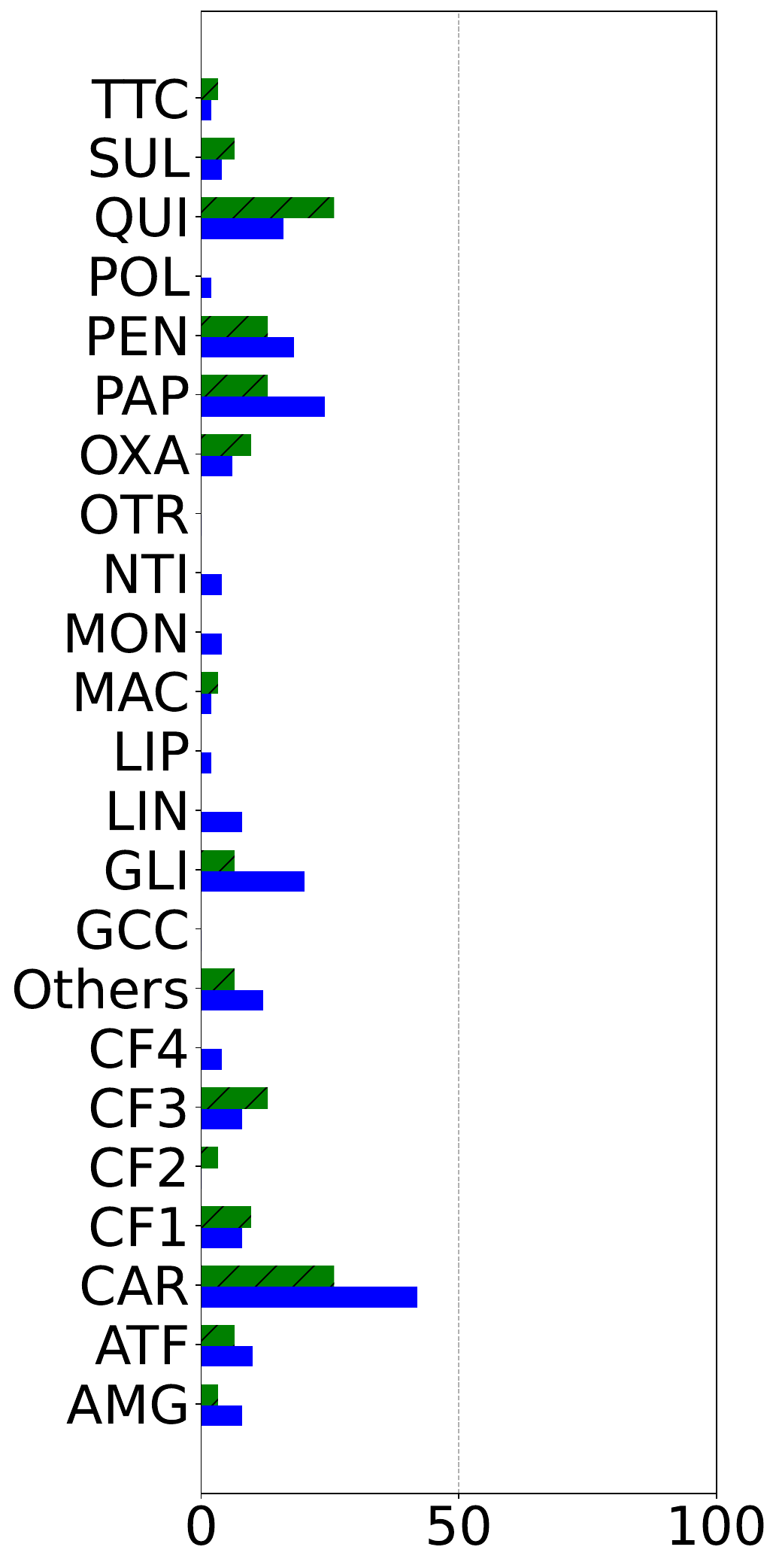}
	\end{subfigure}
	\begin{subfigure}[]
		\centering
         \includegraphics[width=0.18\textwidth]{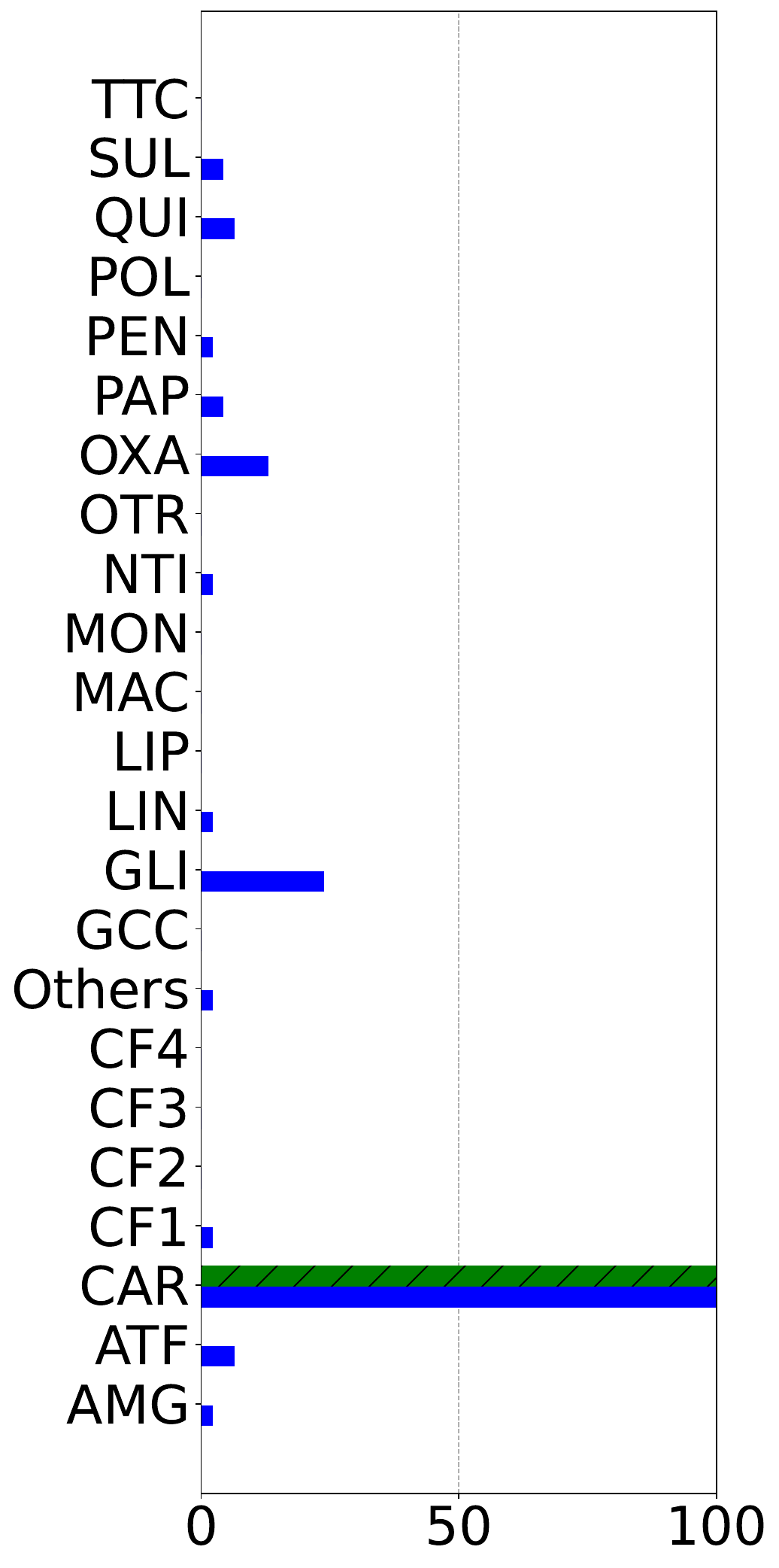}
	\end{subfigure}

    \caption{\textcolor{black}{Each panel corresponds to patients in distinct clusters generated by employing different TS methods, specifically DTW (first row) and TCK (second row): (a) $C_1$; (b) $C_2$; (c) $C_3$; (d) $C_4$; (e) $C_5$. In each figure, the corresponding panel offers the percentages of patients categorized by the antibiotic family they received, with solid blue horizontal bars indicating patients with MDR and green diagonal-striped bars indicating patients without MDR.}}
	\label{vis_4}
\end{figure}

Following the cluster composition analysis based on whether patients exhibited MDR, the next step involves characterizing each cluster by integrating information from the input MTS values corresponding to each patient. To accomplish this, graphical representations were generated (see Figure~\ref{vis_4}), illustrating the percentage of patients belonging to each of the 23 antibiotic groups within each cluster. This was done separately for MDR and non-MDR patients. Figure~\ref{vis_4} presents this information for both MTS methods: DTW and TCK, represented in the first and second rows, respectively. Analyzing the DTW results (first row) reveals no clear grouping based on antibiotics, except for cluster $C_3$, where approximately 70\% of non-MDR patients are taking CF3 antibiotics (see Figure~\ref{vis_4}~(b)). A similar pattern is observed in patients comprising cluster $C_5$ in the case of TCK (see Figure~\ref{vis_4}~(f)). However, in the TCK representation, it is evident that all patients in cluster $C_5$ (Figure~\ref{vis_4}~(i)) are undergoing treatment with CAR antibiotics. Therefore, CF3 and CAR antibiotics may be considered clinically relevant data, indicating a higher prevalence in their administration within each analyzed cluster.

Secondly, we analyzed the percentage of patients requiring MV and the percentage of MDR co-patients for each cluster. The results of this analysis are presented in Table~\ref{Tabla:vis}, both for DTW and TCK, in the first and second rows, respectively. It is noteworthy that for both DTW (cluster $C_2$) and TCK (cluster $C_2$), the majority of patients exhibited a high need for MV, with over 80\% requiring such treatment. Additionally, more than 18\% of these patients shared the ICU with MDR patients. Conversely, patients in cluster $C_4$ (in both DTW and TCK) required less MV treatment. Interestingly, patients in $C_5$ (TCK) did not conform to any of the aforementioned patterns. Despite requiring less MV, they shared the ICU with a higher percentage of MDR co-patients compared to other clusters. This trend could potentially contribute to the development and transmission of multi-resistant strains, exacerbating the problem of antimicrobial resistance in other patients and deteriorating their overall health status.

\begin{table}[h!]
    \caption{Percentage of patients with MV and percentage of MDR co-patients for each cluster.}
    \label{Tabla:vis}
    \centering
    \begin{tabular*}{\textwidth}{@{\extracolsep\fill}c|c|ccccc}
    \toprule
    Method & Variable & $C_1$ &  $C_2$ &  $C_3$ & $C_4$ & $C_5$ \\
    \midrule
    \multirow{2}{*}{DTW} & MV & 69.58\% & 65.16\% & 86.86\% & 51.30\%  & 40.74\% \\
    & MDR co-patients &  15.46\%  & 17.30\% & 18.93\% &  15.79\% & 18.74\% \\
    \midrule
    \multirow{2}{*}{TCK} & MV & 66.35\% & 82.82\% & 53.73\% & 61.70\%  & 41.97\% \\
    & MDR co-patients & 15.73\%  & 18.31\% & 14.59\% &  20.77\% & 18.82\%  \\
    \bottomrule
    \end{tabular*}
\end{table}

Finally, we conducted an analysis of the distribution of values for the most relevant static variables for the patients within each of the five clusters, aiming to identify similarities and differences across clusters. It is essential to emphasize that the static variables were not utilized in creating the predictive models or for clustering. Instead, our objective here is to determine whether the clustering methods based on MTS yield outputs that exhibit any correlation with certain static variables of the patients. As advised by our clinical experts, we examined four static variables: the categorical variables ``origin'' and ``destination'', as well as the continuous variables ``SAPS III Score''~\cite{nassar2014evaluation} and ``age''.

\begin{figure}[h!]
\centering
    \begin{subfigure}[]
		\centering
        \includegraphics[width=0.485\textwidth]{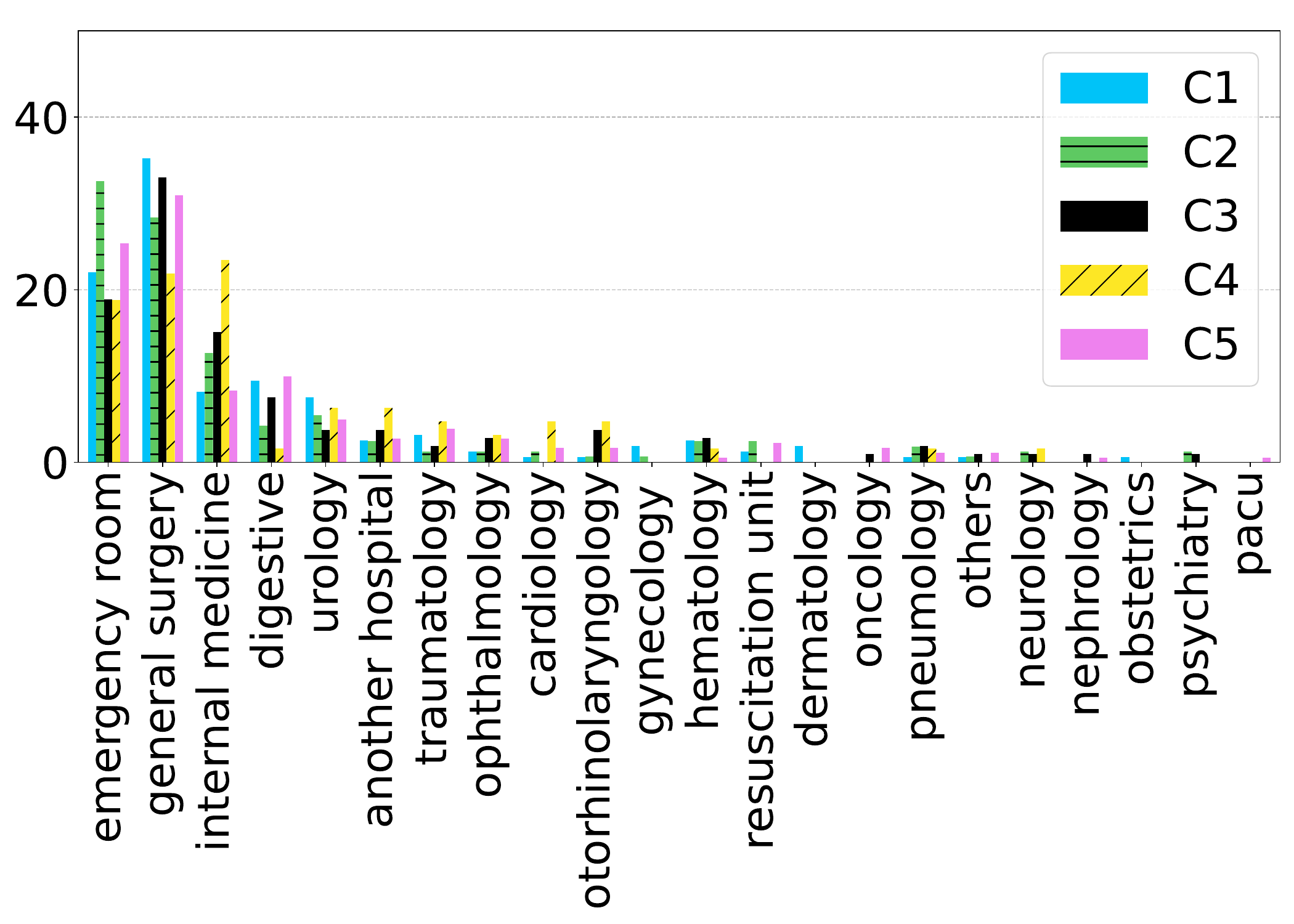}
	\end{subfigure}
	\begin{subfigure}[]
		\centering
        \includegraphics[width=0.485\textwidth]{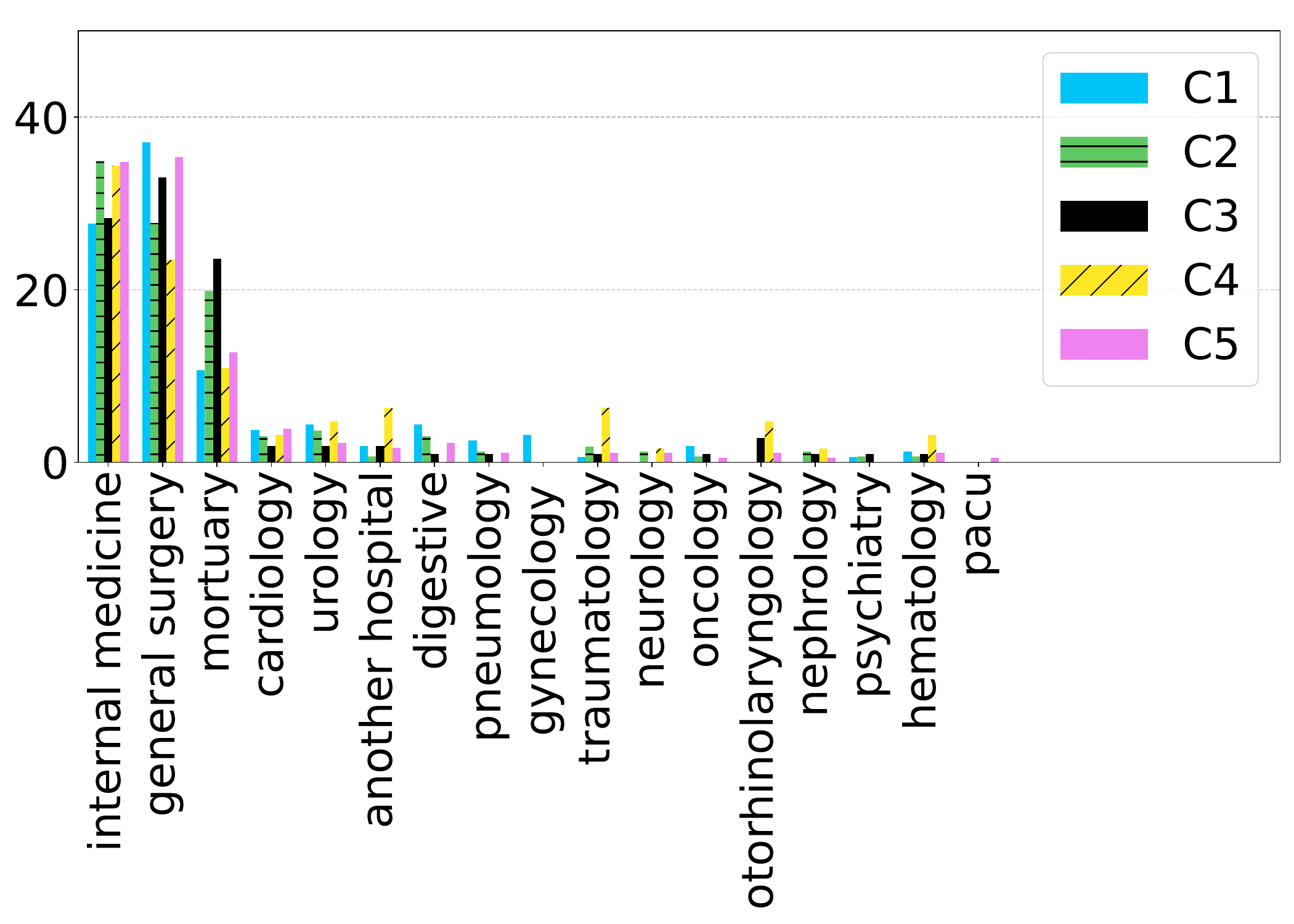}
	\end{subfigure}

 \begin{subfigure}[]
		\centering
        \includegraphics[width=0.485\textwidth]{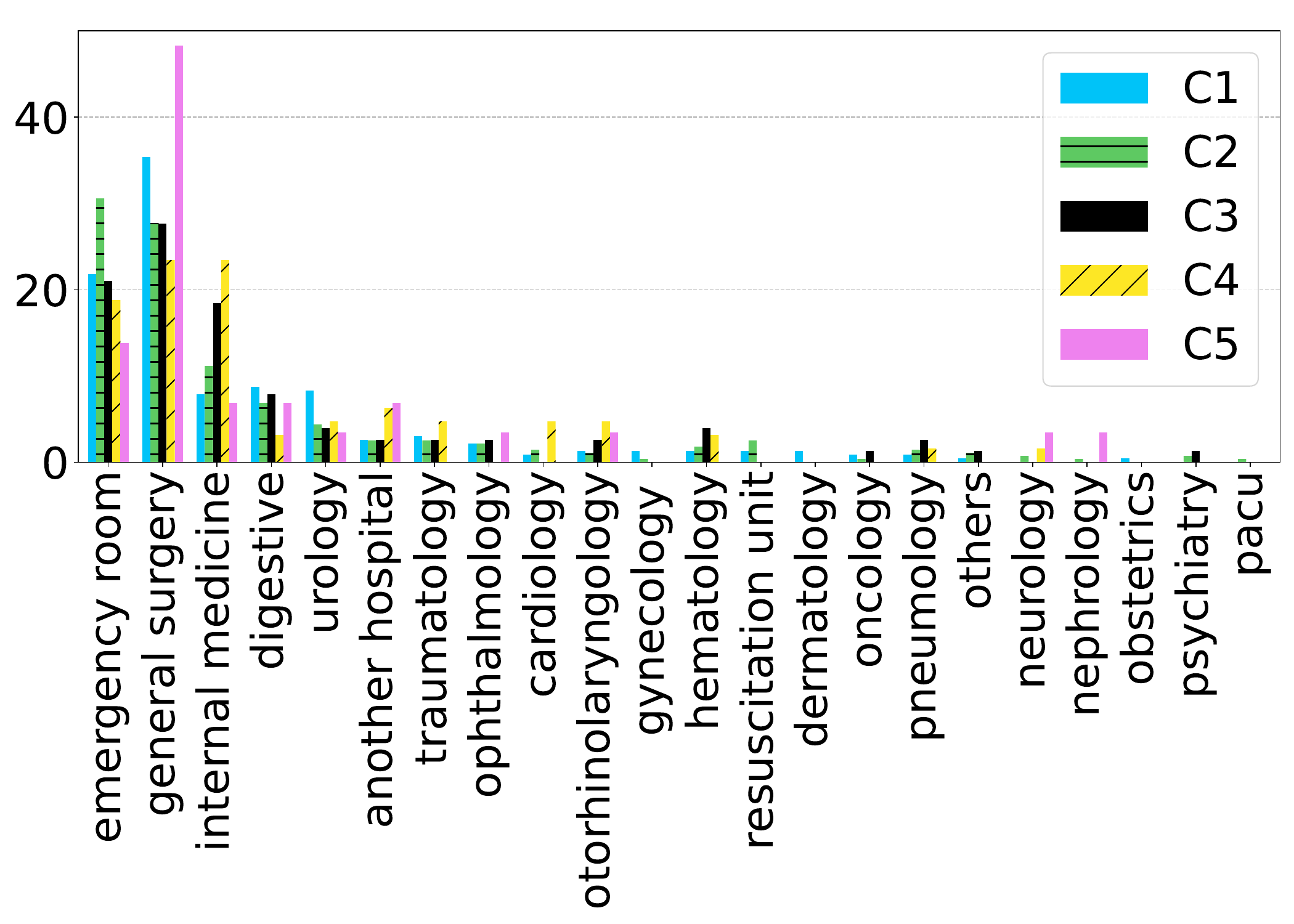}
	\end{subfigure}
	\begin{subfigure}[]
		\centering
        \includegraphics[width=0.485\textwidth]{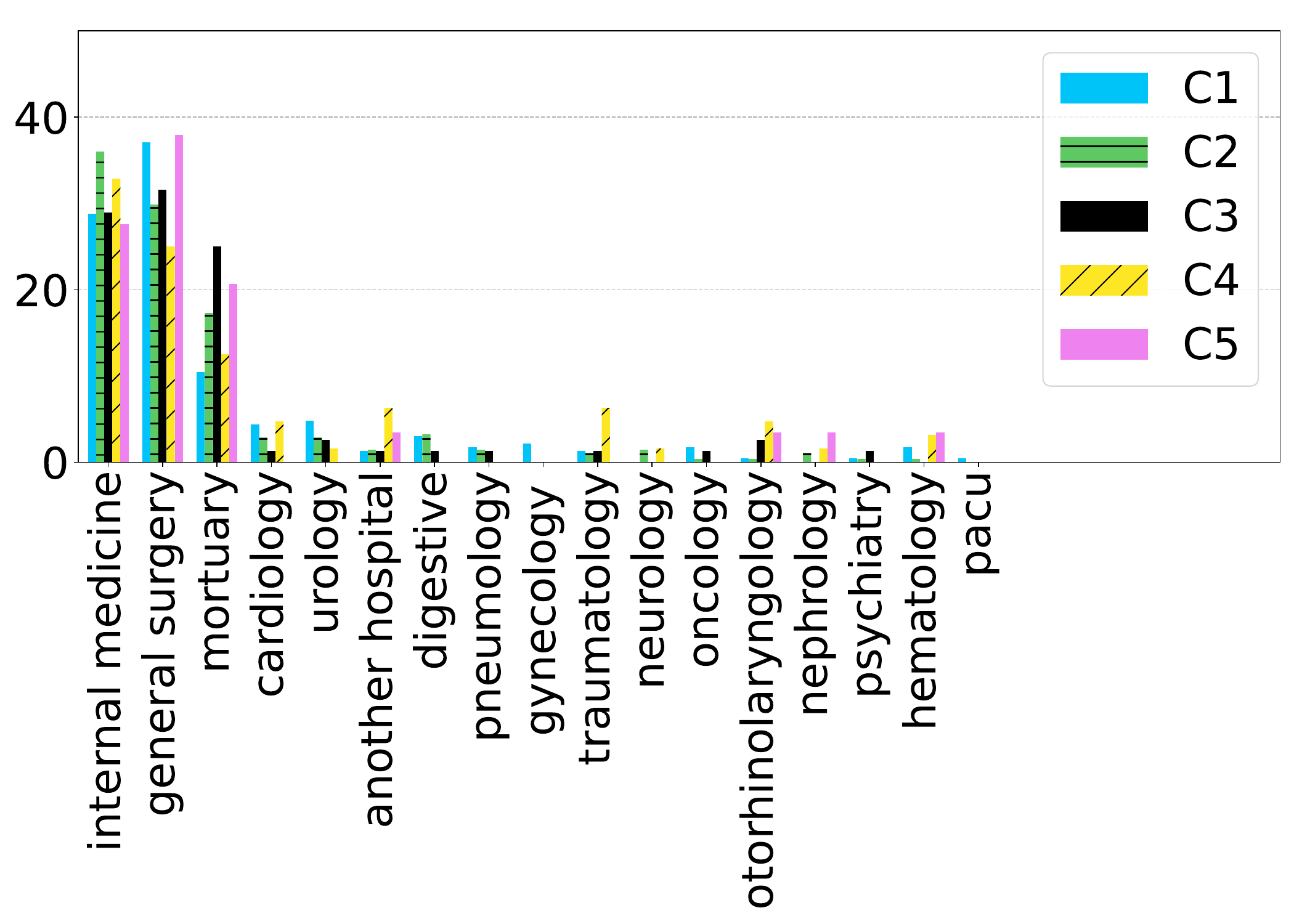}
	\end{subfigure}
        \vspace{-0.3cm}
	\caption{We depict the patient distribution based on their origin (pre-ICU admission) and their destination (post-ICU discharge). The first column displays the point of origin, while the second column represents the destination. The data in the first row pertain to DTW, while the data in the second row pertain to TCK.}
	\label{vis_5}
\end{figure}

We begin with the distribution for the \textit{categorical variables}. It's worth mentioning that the variable ``origin'' indicates the service from which the patient arrived before admission to the ICU, such as the emergency department or internal medicine. Meanwhile, the variable ``destination'' indicates the clinical unit to which the patient was transferred upon leaving the ICU, such as internal medicine, the mortuary, general surgery, etc. The empirical categorical distributions for each of the five clusters are displayed in the two panels of Figure~\ref{vis_5}, based on DTW (first row) and TCK (second row). We observe that, even though the variable ``origin" was not considered for identifying the clusters, distinct variations in patient origins are evident among the clusters, especially in the case of TCK (see Figure~\ref{vis_5}~(c)). Approximately 50\% of the patients in $C_5$ originate from the general surgery department, while patients in the remaining groups predominantly come from the general surgery department. The ``destination'' variable reveals that most patients are transferred to internal medicine and general surgery units, with a higher percentage of deaths in clusters with a greater percentage of patients with MDR: $C_3$ for DTW with 25\%, and $C_3$ and $C_5$ for TCK with 25\% and 20\%, respectively. Of particular interest is $C_1$ for DTW, where non-MDR patients predominate but still exhibit a 20\% mortality rate.

\begin{figure}[h!]
\centering
	\begin{subfigure}[]
		\centering
        \includegraphics[width=0.485\textwidth]{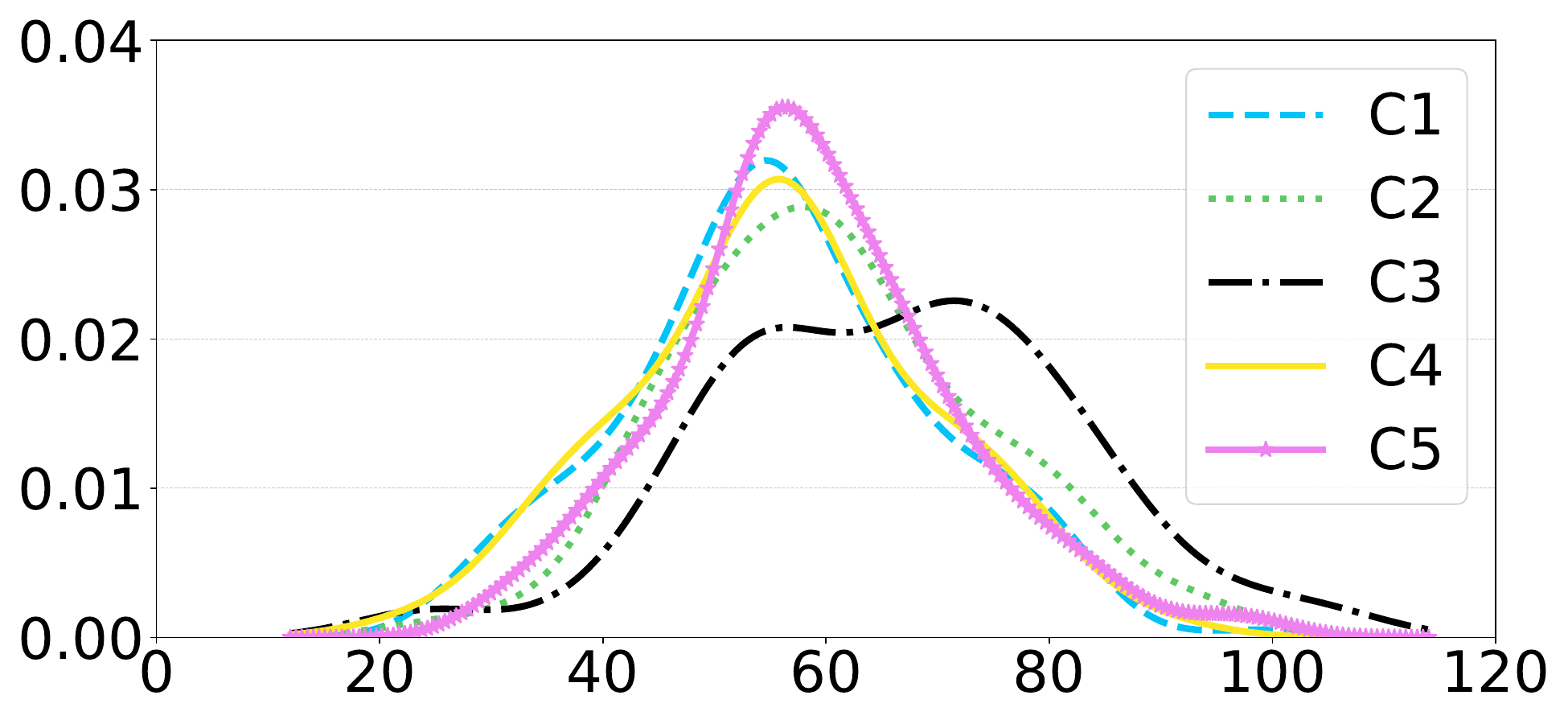}
	\end{subfigure}
	\begin{subfigure}[]
		\centering
        \includegraphics[width=0.485\textwidth]{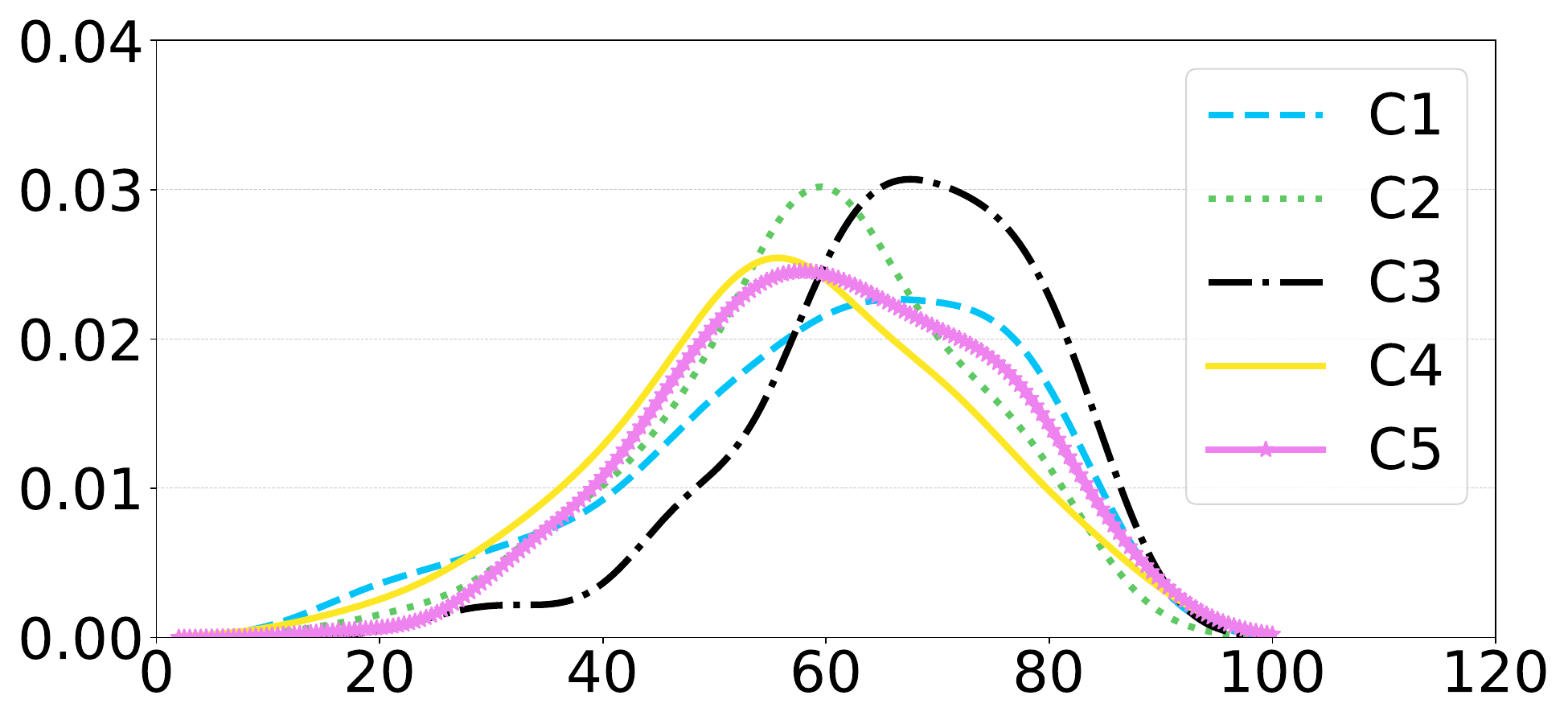}
	\end{subfigure}

 \begin{subfigure}[]
		\centering
        \includegraphics[width=0.485\textwidth]{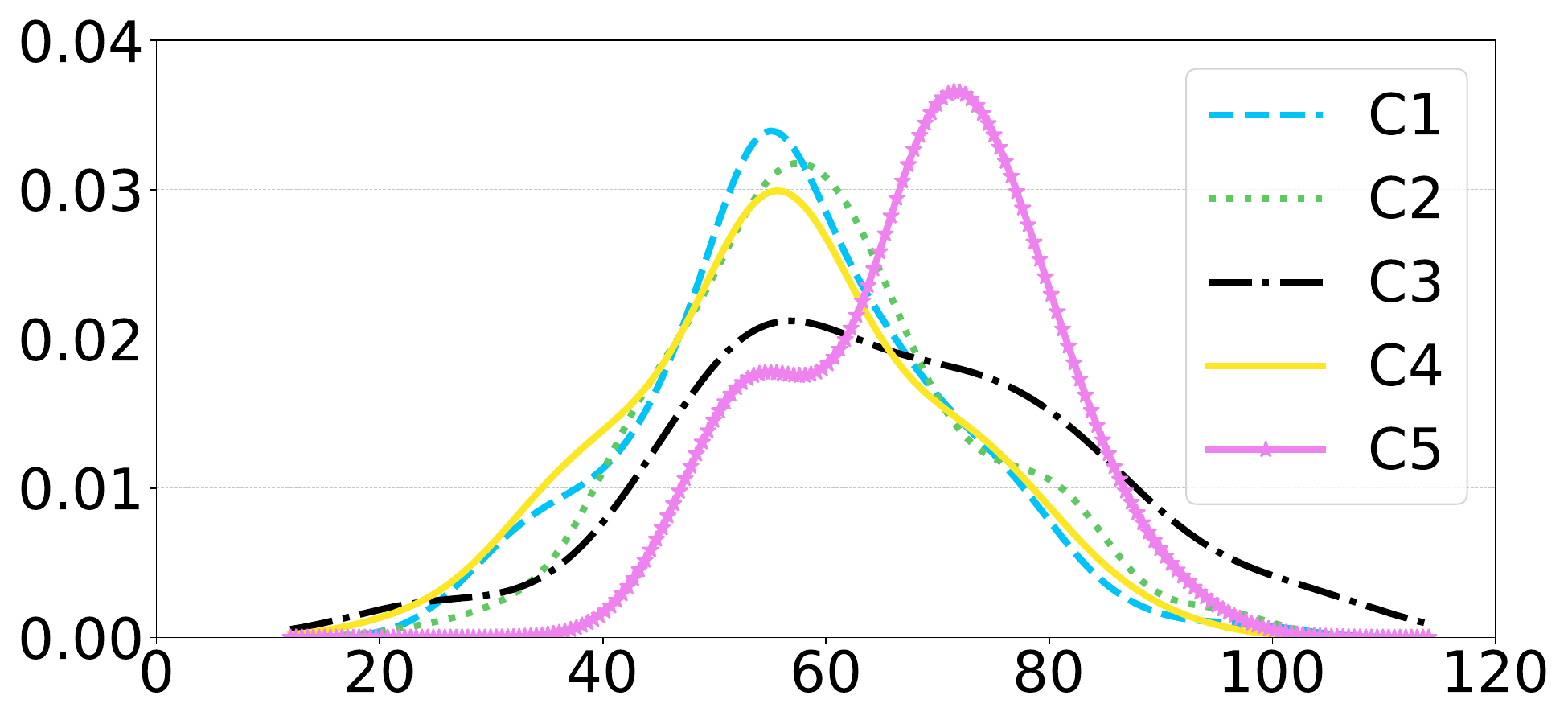}
	\end{subfigure}
	\begin{subfigure}[]
		\centering
        \includegraphics[width=0.485\textwidth]{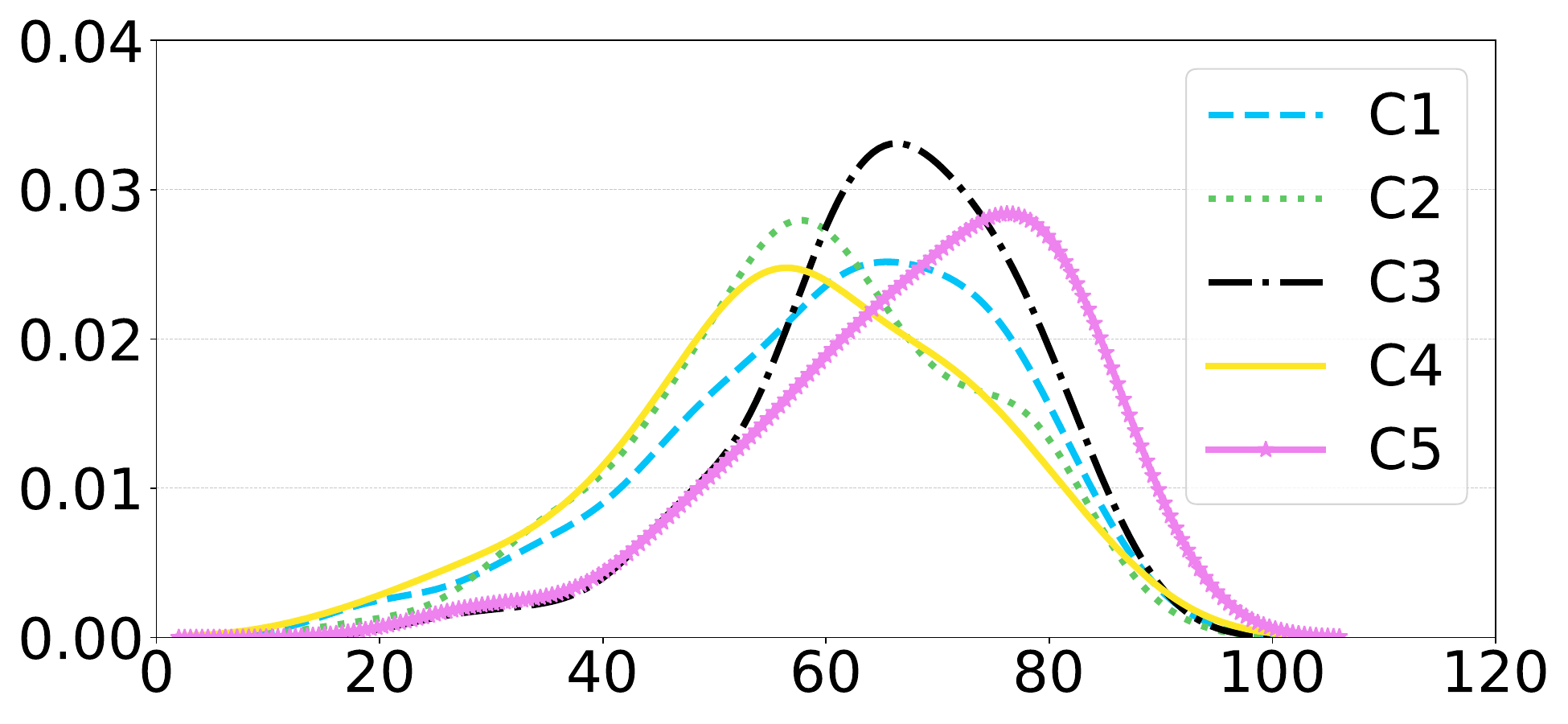}
	\end{subfigure}
        \vspace{-0.3cm}
	\caption{Estimated PDF for SAPS III Score (a)(c), and age (b)(d) in each cluster, considering DTW (first row) and TCK (second row).}
	\label{vis_6}
\end{figure}

Next, we carried out a similar per-cluster analysis, but considering the variables ``SAPS III Score'' and ``age''. Since these two static variables take \textit{continuous values}, instead of bars representing the percentage of each categorical value, we estimated and reported the Probability Density Function (PDF) associated with each variable and cluster (see Figure~\ref{vis_6}). The PDF was estimated using a non-parametric Parzen window technique with Gaussian kernels~\cite{Parzen}, with the width parameter set according to Silverman’s rule~\cite{silverman2018density}. Figure~\ref{vis_6}(a) shows the estimated PDF for ``SAPS III Score'', while Figure~\ref{vis_6}(b) shows the estimated PDF for ``age''. 
An intriguing observation arises from the PDF analysis of ``SAPS III Score'' for patients in clusters $C_3$ (DTW) and $C_5$ and $C_3$ (TCK), as depicted in Figure~\ref{vis_6}(a). Specifically, upon examining the central points of each peak (bell curve), we note that the position of the lower peak in the mentioned clusters coincides with the position of the largest peak (i.e., the mode) in the PDFs of the remaining groups. Conversely, the position of the largest peak in the identified clusters (i.e., the mode for the patients comprising those groups) is shifted to the right, indicating that the majority of patients exhibit a more severe health condition. It is worth emphasizing again that the clustering algorithm had no prior knowledge of the severity score values when forming these clusters.

Turning our attention to the PDFs associated with age, a striking observation is the significantly higher mean and modal values for clusters $C_3$ and $C_1$ (DTW) and $C_5$ (TCK). Specifically, we observe that both the mean and modal ages of patients in $C_2$ and $C_4$ for DTW are 10/15 years higher than those of patients in the other groups. Remarkably, with the TCK method, we can identify a cluster ($C_5$) characterized by elderly patients, with an average age of around 80 years. Thus, these findings suggest that with both methods, we can identify: i) patients in $C_3$ (DTW) and $C_5$ (TCK), where nearly all patients have MDR, consist of older patients with higher SAPS III Scores; and ii) this information was not explicitly supplied to our data analysis process but was effectively extracted by our embedding and clustering approach. However, the TCK method enables a more nuanced characterization of cluster $C_5$ as comprising MDR patients from the general surgery department, all treated with CAR antibiotics, who are of advanced age and in more critical health conditions.

\section{Discussion}
\label{sec:Disc}



\textcolor{black}{This work introduces a novel architecture for modeling MTS and heterogeneous clinical data to predict MDR in ICU patients. The growing prevalence of MDR, often driven by antibiotic misuse, represents a major threat to healthcare systems—particularly in intensive care settings, where patient vulnerability is heightened by clinical severity and invasive procedures~\cite{de2018antimicrobial}. Addressing this challenge requires predictive models that are not only accurate but also interpretable and robust to real-world clinical variability.}

\textcolor{black}{To meet this need, we propose a framework that integrates MTS analysis with distance-based learning and kernel transformations, leveraging DTW and TCK to construct patient-to-patient similarity matrices. These representations enable reliable MDR prediction while supporting interpretability through the identification of temporal risk patterns and clinically coherent patient clusters. Building on this foundation, the incorporation of dimensionality reduction and kernel-based clustering further allows the construction of graph-based patient representations that capture latent relationships in the data. This enhances both model performance and transparency.}

\textcolor{black}{We validated the proposed framework using real-world ICU data from the UHF, achieving an ROC-AUC of 81\%, which outperforms several state-of-the-art ML and DL baselines evaluated on the same dataset. Notably, this improvement is achieved without compromising interpretability, a frequent limitation of complex DL models. Compared to prior studies reporting ROC-AUC values between 66\% and 77\%~\cite{rich2022development, rhodes2023machine}, our model demonstrates superior predictive power by combining DTW or TCK with exponential kernels and a $\nu$-SVM. These configurations consistently revealed five clinically meaningful patient clusters. For instance, MDR prevalence was concentrated in cluster $C_3$ (DTW) and clusters $C_3$–$C_5$ (TCK), while cluster $C_2$ (DTW) was linked to non-MDR patients through feature CF3. Cluster $C_5$ (TCK) emphasized the role of invasive procedures and MDR co-patient exposure. These findings highlight the framework's capacity to extract actionable insights from MTS data.}

\textcolor{black}{Despite these promising results, several limitations must be acknowledged. The proposed framework relies on access to temporally rich patient records, which may limit its immediate applicability in real-time settings where data accrues progressively throughout a patient's ICU stay. Additionally, the model was trained on data from a single institution, which may restrict its generalizability to other hospitals due to differences in population profiles, screening protocols, and EHR systems. Direct transfer of trained models across institutions may degrade performance due to domain shifts. To address these challenges, future work should explore transfer learning strategies—such as fine-tuning with local patient data, meta-learning, or domain adaptation techniques—to improve model adaptability across diverse clinical environments.}

\textcolor{black}{In light of these considerations, external validation across multiple hospital systems and patient cohorts will be essential to confirm the robustness of our approach. While our study ensured balanced representation across different MDR screening protocols and consistently applied standardized MDR definitions, the potential for latent biases due to historical changes in clinical practices remains. Proactive mitigation of these biases is critical for fair and reliable model deployment.}

\textcolor{black}{Looking ahead, the predictive power of this framework could be further enhanced by incorporating additional clinical variables such as oxygen saturation, temperature trends, heart rate variability, inflammatory markers, comorbidities, and prior antibiotic exposure. Expanding the model to larger and more diverse datasets will also be vital for broader applicability. To refine feature relevance and improve representation learning, future work could integrate multimodal feature selection techniques—including Laplacian-based methods~\cite{khachatrian2021multimodal} and hypergraph-based approaches~\cite{shao2020hypergraph}—alongside advanced autoencoder architectures such as variational~\cite{lin2020deep}, stacked sparse~\cite{zhang2021pseudo}, and adversarial models~\cite{hara2020intrusion}.}

\textcolor{black}{In conclusion, this framework advances clinical decision support by bridging the critical temporal gap inherent in conventional microbiological diagnostics, which typically require 48–72 hours for confirmation. By enabling early, interpretable predictions before laboratory results, the model empowers clinicians to initiate targeted treatments or isolation protocols proactively. Its interpretability further supports evidence-based decision-making by identifying high-risk profiles—such as prolonged ICU stays, broad-spectrum antibiotic use, and MDR co-exposure—even in the absence of microbiological confirmation. Altogether, this dual focus on accuracy and interpretability transforms MDR management from reactive to proactive, potentially reducing transmission, optimizing treatment strategies, and improving ICU outcomes.}

\section*{Declaration of competing interest}

The authors declare that they have no known competing financial interests or personal relationships that could have appeared to influence the work reported in this paper.

\section*{Acknowledgments}
\label{acks}

This work was supported by the Spanish AEI (DOI 10.13039/501100011033) under Grants PID2022-136887NBI00 and PID2023-149457OB-I00, and by the Autonomous Community of Madrid within the ELLIS Unit Madrid framework and Grant TEC-2024/COM-89.

\bibliographystyle{elsarticle-num} 
\bibliography{referencesclean}





\end{document}